\documentclass{article}

\usepackage{iclr2027_conference,times}
\usepackage{amsmath,amsfonts,bm}

\def\eqref#1{equation~\ref{#1}}
\def\1{\bm{1}}

\DeclareMathAlphabet{\mathsfit}{\encodingdefault}{\sfdefault}{m}{sl}
\SetMathAlphabet{\mathsfit}{bold}{\encodingdefault}{\sfdefault}{bx}{n}

\newcommand{\HieraPredDim}{36}
\newcommand{\HieraScaleDecodedDim}{24}
\newcommand{\HieraFullDecodedDim}{4}

\newcommand{\HieraBaseDim}{64}
\newcommand{\HieraTemporalOps}{13}

\newcommand{\HieraGainU}{12.0\%}
\newcommand{\HieraGainM}{15.5\%}

\usepackage{amsmath}
\usepackage{amssymb}
\usepackage{booktabs}
\usepackage{graphicx}
\usepackage{multirow}
\usepackage{array}
\usepackage{tabularx}
\usepackage{colortbl}
\usepackage{flafter}
\usepackage{float}
\usepackage{placeins}
\usepackage{hyperref}
\usepackage{url}


\title{HieraNTP: Structured Risk Representations for Time-Series Anomaly Detection via Hierarchical Next-Token Prediction}

\author{%
{\small
\textbf{Yujie Li}$^{1,2}$,\kern1.25pt
\textbf{Zezhi Shao}$^{1}$\thanks{Corresponding authors: Zezhi Shao and Fei Wang.}\kern0.4em,\kern1.25pt
\textbf{Boyu Diao}$^{1,2}$,\kern1.25pt
\textbf{Zhulin An}$^{1,2}$,\kern1.25pt
\textbf{Yongjun Xu}$^{1,2}$,\kern1.25pt
\textbf{Xueqi Cheng}$^{1,2}$,\kern1.25pt
\textbf{Fei Wang}$^{1,2,*}$} \\
\normalfont $^{1}$State Key Laboratory of AI Safety, Institute of Computing Technology, \\
\normalfont Chinese Academy of Sciences, Beijing, China \\
\normalfont $^{2}$University of Chinese Academy of Sciences, Beijing, China \\
\normalfont\texttt{\{liyujie23s,shaozezhi,diaoboyu2012,anzhulin\}@ict.ac.cn} \\
\normalfont\texttt{\{xyj,cxq,wangfei\}@ict.ac.cn}
}

\hypersetup{
  hidelinks,
  pdftitle={HieraNTP: Structured Risk Representations for Time-Series Anomaly Detection via Hierarchical Next-Token Prediction},
  pdfauthor={Yujie Li, Zezhi Shao, Boyu Diao, Zhulin An, Yongjun Xu, Xueqi Cheng, Fei Wang}
}

\iclrfinalcopy

\begin{document}
\raggedbottom

\maketitle
% Preprint: clear the conference-status header set by the template.
\fancyhead{}

\begin{abstract}
Time-series anomaly detection commonly converts deviations from learned
temporal regularities into anomaly scores. This practice can conflate
violations of temporal regularity with task-specific anomaly judgments:
the same deviation may be anomalous under one task definition and benign
under another. Aggregating heterogeneous deviations into a single score
before task-specific adaptation can obscure the evidence needed to
distinguish these cases. We introduce \textbf{HieraNTP}, a
\textbf{Hierarchical Next-Token Prediction} framework that separates
self-supervised predictive modeling from task-specific risk interpretation.
HieraNTP pretrains a hierarchical tokenizer and a causal next-token predictor
to capture temporal regularities at multiple scales. It then organizes
prediction discrepancies into a structured risk representation, preserving
complementary evidence across temporal scales in token, local-statistic,
and observation spaces. A task-adaptive detector uses anomaly labels to
select and combine this evidence for the target task.
Across anomaly detection
benchmarks, HieraNTP improves VUS-PR by \textbf{\HieraGainU} on TSB-AD-U and
\textbf{\HieraGainM} on TSB-AD-M relative to the strongest baselines,
and achieves the highest composite score on both TAB tracks, with controlled
comparisons supporting the value of retaining distinct evidence.
\mbox{Our code is available at \url{https://github.com/blisky-li/HieraNTP}.}
\end{abstract}

\section{Introduction}
\label{sec:introduction}

Time-series anomaly detection (TSAD) identifies observations or intervals that
depart from expected temporal behavior. It supports continuous
monitoring of cloud services, industrial systems, and cyber-physical
infrastructure, where anomalous events may signal service failures, equipment
faults, or attacks
\citep{xu2018donut,su2019omnianomaly,tuli2022tranad,khan2026chronosad}.

Existing TSAD methods commonly learn temporal regularities and detect
anomalies directly from forecasting or reconstruction errors
\citep{xu2018donut,su2019omnianomaly,audibert2020usad,tuli2022tranad,liu2024tsbad,pmlr-v235-sarfraz24a}.
Association- and contrastive-learning methods measure inconsistencies in
temporal dependencies or representations
\citep{xu2022anomalytransformer,yang2023dcdetector,lan2025timercd,han2025timerep}.
Temporal pretraining learns transferable forecasting and representation capabilities
\citep{pmlr-v235-goswami24a,ansari2024chronos,pmlr-v235-das24c,han2025timerep,khan2026chronosad}.
Larger departures generally indicate stronger anomaly evidence.

\begin{figure}[!t]
    \centering
    \includegraphics[width=0.9\textwidth]{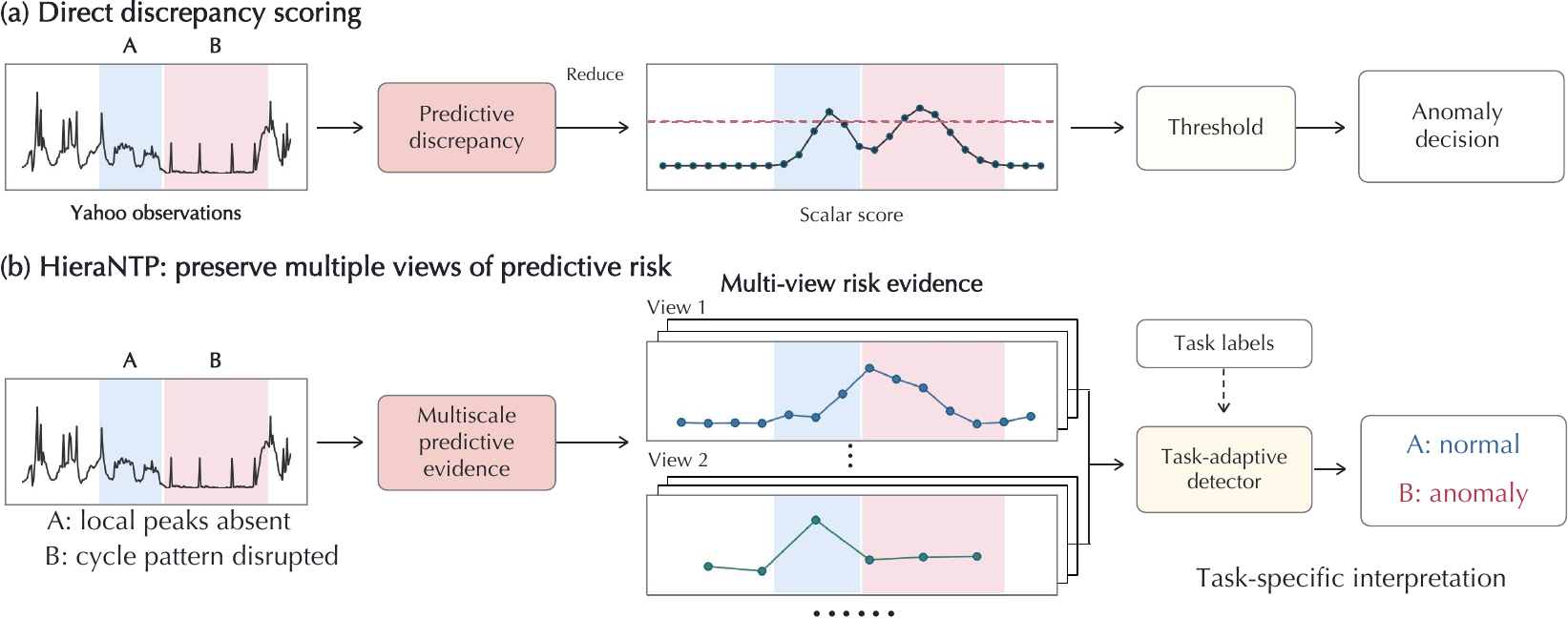}
    \caption{Schematic motivation using a Yahoo sequence from TSB-AD.
    The task treats missing peaks (A)
    as normal and disrupted cycles (B) as anomalous.
    (a) Early scalar reduction can obscure this distinction.
    (b) HieraNTP retains a structured risk representation for task-adaptive detection.}
    \label{fig:motivation}
\end{figure}

Direct discrepancy scoring can conflate violation of learned regularity
with task-defined anomaly. The same deviation may be anomalous
under one task definition and benign under another.
Error magnitude alone may not capture the form and context relevant to these judgments.
Contextual anomalies can disrupt temporal relations with small value deviations
\citep{lan2025timercd}, while persistent abnormal dynamics may become
predictable within context, reducing forecasting errors \citep{uray2026exploring}.
Forecasting or reconstruction gains may yield little improvement in detection,
and simple local statistics can closely approximate foundation-model errors
\citep{zhu2026oneliners}. We refer to this mismatch between
task-agnostic predictive evidence and task-specific anomaly interpretation as
the \emph{predictive risk--decision gap}.

Task-specific interpretation depends on the information retained in predictive
evidence. Figure~\ref{fig:motivation} shows two types of deviation within one
task: missing local peaks in region A and disrupted cycles in region B.
Although both violate the preceding pattern, the illustrated task labels only
region B as anomalous.
Distinguishing them requires evidence about how each departs from that pattern.
Scalar reduction can discard these distinctions before a decision boundary is
learned, motivating the retention of complementary discrepancies across spaces
and temporal supports.

We introduce \textbf{HieraNTP}, a \textbf{Hierarchical Next-Token Prediction}
framework that separates self-supervised predictive modeling from
task-specific risk interpretation. Its multiscale tokenizer produces continuous
causal tokens, and its hierarchical predictor models their evolution.
Comparing predicted and realized targets yields token-space,
local-statistic, and observation-space discrepancies. Temporal summaries preserve
sustained and peak responses, while variable aggregation captures broad or
concentrated departures. With the tokenizer and predictor frozen, target labels
train the task-adaptive detector on the resulting structured risk representation.

On the official TSB-AD benchmark \citep{liu2024tsbad}, HieraNTP outperforms the
strongest baseline in VUS-PR on both the univariate and multivariate tracks.
HieraNTP also achieves the highest dataset-mean VUS-PR on both TAB tracks
\citep{qiu2025tab}.
Our contributions are:
\begin{itemize}
  \item We identify the predictive risk--decision gap and explain how scalar
  scoring can discard distinctions needed for task-specific anomaly decisions.
  \item We propose HieraNTP, which converts multiscale next-token prediction
  discrepancies into a structured risk representation before task-specific
  anomaly scoring.
  \item Under matched detector training, multiview evidence outperforms scalar
  time-domain prediction errors and mean, maximum, or mean--maximum aggregation on TSB-AD.
\end{itemize}

\section{Related Work}
\label{sec:related}

\noindent\textbf{Deviation-based time-series anomaly detection.}
Similarity-based detectors identify subsequences that lack
close historical matches \citep{yeh2016matrixprofile}. Generative detectors
measure reconstruction discrepancies or likelihoods under learned normal
dynamics
\citep{xu2018donut,su2019omnianomaly,audibert2020usad,tuli2022tranad,fang2024tfmae},
whereas forecasting-based detectors compare predicted and realized
observations \citep{liu2024tsbad,pmlr-v235-sarfraz24a}. Association and
contrastive approaches instead construct anomaly scores from inconsistencies
in temporal dependencies or learned representations
\citep{xu2022anomalytransformer,yang2023dcdetector}.

\noindent\textbf{Temporal pretraining and next-token prediction.}
Pretrained time-series models acquire transferable temporal structure through
reconstruction or forecasting objectives \citep{shao2025blast}. MOMENT uses masked reconstruction,
Chronos performs autoregressive prediction over discretized values, and
TimesFM forecasts continuous patches
\citep{pmlr-v235-goswami24a,ansari2024chronos,pmlr-v235-das24c}. Controlled
evaluations find that their reconstruction or forecasting scores can resemble
simple local statistics and that improved prediction quality can leave
anomaly rankings unchanged \citep{zhu2026oneliners}.
Persistent abnormal dynamics can become predictable within context,
concentrating forecasting discrepancies near event boundaries
\citep{uray2026exploring}.

\noindent\textbf{Anomaly-oriented pretraining and task adaptation.}
Anomaly-oriented models introduce detection-specific information at different
stages of learning. DADA uses explicit normal--anomalous differentiation during
pretraining \citep{shentu2025dada}. TimeRCD learns relative-context discrepancy
from synthetic sequences with context-dependent anomaly labels and applies
the resulting detector to unseen datasets in zero-shot evaluation
\citep{lan2025timercd}. TSPulse combines reconstruction and prediction
scores from multiple heads through fixed statistics, or selects a head using
labeled tuning data. Its target-data fine-tuning uses self-supervised
objectives \citep{ekambaram2025tspulse}. Other adaptations retain a pretrained
temporal model and construct anomaly scores from intermediate representation
distances, supervised classifiers, or calibrated forecasting residuals
\citep{han2025timerep,khan2026chronosad,gil2026adaptive}.

HieraNTP retains multiscale discrepancies from hierarchical next-token prediction,
comparing predicted and realized tokens, local statistics
\citep{wang2025aries,li2026apt}, and decoded time-series segments.
A task-adaptive detector learns joint decision rules over this structured risk
representation from target labels, with the tokenizer and predictor frozen.

\section{Method}
\label{sec:method}

\subsection{Problem Formulation and Framework}

\begin{figure}[!t]
  \centering
  \includegraphics[width=0.85\textwidth]{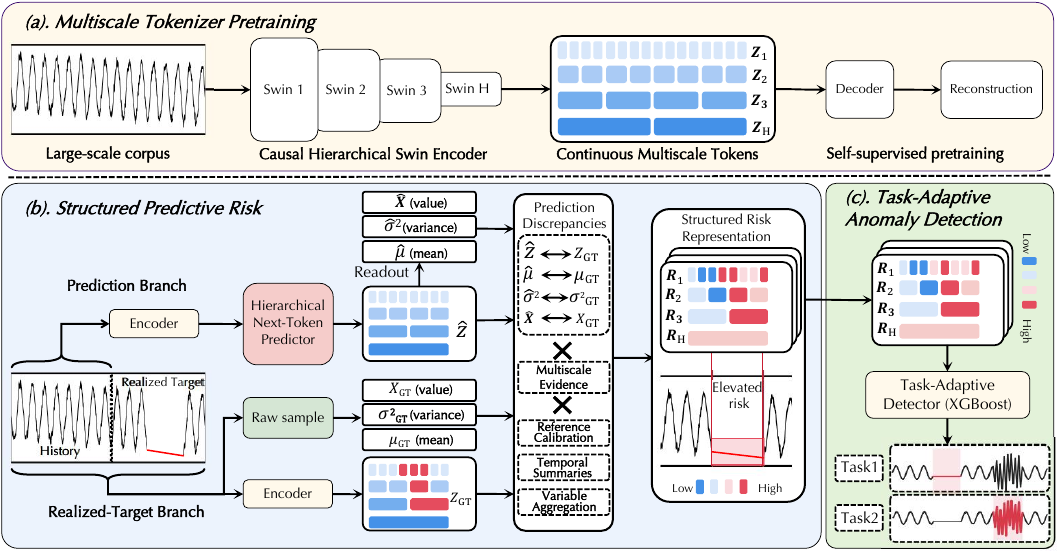}
  \caption{Overview of HieraNTP. (a) Multiscale reconstruction pretrains the
  causal tokenizer. (b) Prediction and realized-target ($\mathrm{GT}$) branches
  yield token-, local-statistic-, and
  observation-space discrepancies. (c) Labels train the task-adaptive detector
  with predictive modules frozen.}
  \label{fig:overview}
\end{figure}

Let $X=(x_{t,c})\in\mathbb R^{T\times C}$ denote a time series with $T$
observations and $C$ variables. HieraNTP assigns an anomaly score $s_t$ to each
time index $t$. As shown in Figure~\ref{fig:overview}, a causal multiscale
tokenizer and a hierarchical next-token predictor are pretrained with
self-supervised objectives. Both are then frozen, and their prediction
discrepancies form the structured risk representation $\mathbf R_t$.
Labels $y_t\in\{0,1\}$ supervise a task-adaptive detector that maps
$\mathbf R_t$ to $s_t$.

\subsection{Multiscale Tokenizer Pretraining}
\label{sec:tokenizer-pretraining}

Anomalies range from brief impulses to disrupted cycles, motivating a
hierarchy of temporal tokens for each variable.
For a normalized variable $X_{1:T,c}$, the tokenizer encoding map $E_\theta$,
including the continuous-token projection, produces
\begin{equation}
  Z=E_\theta(X_{1:T,c})=\{Z_h\}_{h=1}^{H},
  \qquad
  Z_h=(\mathbf z_{\tau,h})_{\tau=1}^{N_h},
  \label{eq:multiscale-tokenizer}
\end{equation}
Patch lengths are $\delta_h\in\{8,16,32,64,128,256\}$ observations.
Past-only local attention ensures causality at patch endpoints.
Fine and coarse levels capture local variation and longer patterns.

Continuous tokens are regularized by a shared codebook, and a multiscale
decoder reconstructs the original sequence and the temporal content at each
native scale \citep{liu2021swin,oord2017neural}. Tokenizer pretraining minimizes
\begin{equation}
  \mathcal L_{\mathrm{tok}}
  =\mathcal L_{\mathrm{full\mbox{-}rec}}
  +\lambda_{\mathrm{ms}}\sum_{h=1}^{H}\mathcal L_{\mathrm{rec}}^{(h)}
  +\mathcal L_{\mathrm{VQ}} .
  \label{eq:tokenizer-objective}
\end{equation}
Reconstruction grounds tokens in observable content and supplies prediction
targets at multiple supports. Token targets are the continuous encoder
projections used for codebook lookup. The encoder, codebook, and decoder are
then frozen.

\subsection{Structured Predictive Risk via Hierarchical Next-Token Prediction}
\label{sec:next-token-risk}

\noindent\textbf{Learning regular temporal evolution.}
With the tokenizer frozen, the predictor estimates next tokens and local
statistics from preceding observations. At target position $\tau$,
$\mathcal M_\tau=\{\mathcal M_{\tau,h}\}_{h=1}^{H}$ collects encoder states
preceding the target interval at each hierarchy level. Prediction proceeds
from fine to coarse levels: fine-level hidden states are merged by temporal
overlap into coarser queries, then updated by causal self-attention and
cross-attention to the corresponding historical states. Level-specific heads
estimate
\begin{equation}
  (\widehat{\mathbf z}_{\tau,h},
   \widehat\mu_{\tau,h},
   \widehat\ell_{\tau,h})
  =\mathcal P_{\omega,h}(\mathcal M_\tau),
  \qquad
  \widehat\sigma_{\tau,h}^{2}=\exp(\widehat\ell_{\tau,h}).
  \label{eq:hierarchical-ntp}
\end{equation}
Self-supervised training matches the predicted token, patch mean, and log
variance to targets from the tokenizer and observed patches.

\noindent\textbf{Complementary views of predictive risk.}
The decoder maps predicted and realized tokens to time-series segments.
Token-space discrepancies measure disagreement in the temporal
content represented by predicted and realized tokens. Local-statistic
discrepancies compare predicted and realized patch means and variances to
measure departures in level and dispersion. Observation-space evidence comprises
decoded-observation discrepancies against observed values and decoded-target
discrepancies against reconstructions from realized target tokens.

\noindent\textbf{Organizing predictive evidence.}
The four components in Figure~\ref{fig:overview}(b) organize these discrepancies
into the structured risk representation:
\begin{itemize}
  \item \textbf{Multiscale Evidence.}
  HieraNTP retains temporal content over six patch lengths, with token-grid
  offsets of 0 and 4 observations providing alternative boundaries. Native-scale and
  full-resolution comparisons yield 64 base discrepancy channels per variable.

  \item \textbf{Reference Calibration.}
  Robust statistics from each sequence's designated training prefix
  $X_{\mathrm{ref}}$ express responses relative to their baseline level and
  variability, accounting for differing error scales across sources and
  variables \citep{deng2021graph}.

  \item \textbf{Temporal Summaries.}
  Each evidence sequence retains its current value and centered means and
  maxima over six window lengths, yielding 13 temporal summaries. Means capture
  sustained responses, while maxima preserve local peaks that averaging can
  attenuate. These windows complement the patch lengths defining prediction
  targets by summarizing duration-dependent discrepancy responses
  \citep{zhu2026oneliners}.

  \item \textbf{Variable Aggregation.}
  Variable means summarize broad responses, while maxima retain departures in
  a few variables. Pool-then-calibrate (Pool$\rightarrow$Cal.) calibrates the
  aggregate, while calibrate-then-pool (Cal.$\rightarrow$Pool) calibrates individual
  variables before aggregation. Univariate aggregation is the identity.
\end{itemize}

These components form three parallel processing paths: Direct,
Pool$\rightarrow$Cal., and Cal.$\rightarrow$Pool.
The Direct path projects reference-standardized patch responses over their
target intervals before variable aggregation and within-sequence ranking.
The Pool$\rightarrow$Cal. and Cal.$\rightarrow$Pool paths
hold each response until the next valid update in the processing direction.
Full-resolution discrepancies remain pointwise. All three paths and their
temporal summaries are computed in both directions. Let
$\mathcal E_{t,c}^{(d)}=\{e_{t,c,j}^{(d)}:j=1,\ldots,\HieraBaseDim\}$ denote the
base channels for variable $c$, direction $d$, and emission time $t$. The index
$j$ identifies the source, patch length or full resolution, and grid alignment. Combining
the three processing variants with the 13 temporal summaries gives
\begin{equation}
  \mathbf R_t^{(d)}
  =
  \Phi\!\left(
    \{\mathcal E_{t',c}^{(d)}\}_{t',c};
    X_{\mathrm{ref}}
  \right),
  \label{eq:structured-risk-representation}
\end{equation}
where $t'$ indexes observations and $\Phi$ applies each variant's time alignment
and evidence transformations, followed by temporal summaries.
The representation contains $64\times3\times13=2496$ features for univariate
sequences and 4992 for multivariate sequences with both variable means and
maxima.

\subsection{Task-Adaptive Anomaly Detection}
\label{sec:downstream-detector}

HieraNTP addresses the predictive risk--decision gap by learning evidence
combinations for target anomalies. With the tokenizer, decoder, and predictor
frozen, an XGBoost detector $g_\psi$ \citep{chen2016xgboost} learns nonlinear
decision rules from structured risk representations and anomaly labels.
Each benchmark track fits a shared base detector across its training data.

Dataset-specific extensions continue boosting with the dataset's labels,
adding corrective trees while keeping the base fixed. All trees share the same
risk representation, and their contributions are summed before the sigmoid.
Inference uses the adapted detector when available and the shared base otherwise,
averaging its outputs across both temporal directions at each timestamp.
Complete-sequence ranks and centered temporal summaries define this offline
configuration.

\subsection{Pretraining Procedure}
\label{sec:pretraining}

\noindent\textbf{Multiscale tokenizer pretraining.}
We jointly train the causal encoder, shared codebook, and decoder on the
QUALS-organized GIFT-Eval Pretrain corpus \citep{li2027quals} with
Eq.~(\ref{eq:tokenizer-objective}).
After lower-learning-rate refinement, the frozen tokenizer supplies encoder
states as predictor context and projected continuous tokens as fixed targets.

\noindent\textbf{Hierarchical next-token pretraining.}
Observed history conditions token and patch-statistic prediction at every scale.
Teacher forcing uses encoded observations, whereas autoregressive rollout uses
preceding predictions. Normalization statistics use only the observed history.

To encourage learning of normal temporal regularities, we continue predictor
training with synthetic sequences and gradually increase real data from QUALS
until the real-to-synthetic ratio reaches 4:1.
Appendices~\ref{app:method-details}--\ref{app:implementation} detail evidence
construction, synthetic data generation, and model training.

\section{Experiments}
\label{sec:experiments}

\begin{figure}[!ht]
    \centering
    \begin{minipage}[t]{0.485\textwidth}
        \centering
        \includegraphics[width=\linewidth]{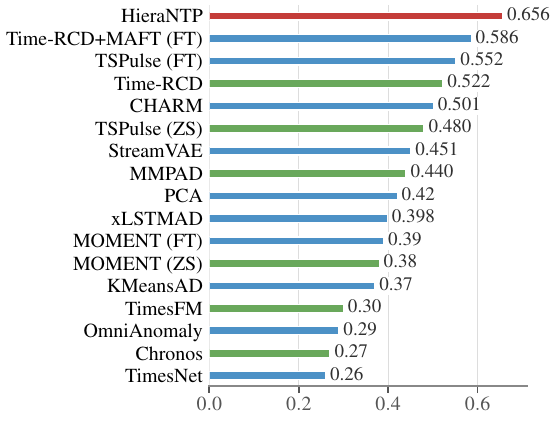}
        \small (a) TSB-AD-U
    \end{minipage}\hfill
    \begin{minipage}[t]{0.485\textwidth}
        \centering
        \includegraphics[width=\linewidth]{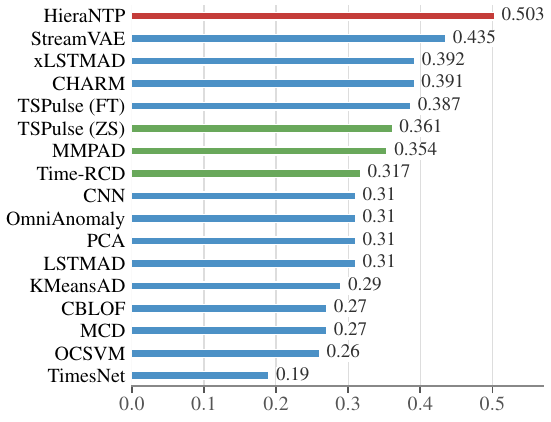}
        \small (b) TSB-AD-M
    \end{minipage}
    \caption{TSB-AD Eval VUS-PR. HieraNTP is red, target-adapted methods are
    blue, and zero-shot or training-free methods are green.}
    \label{fig:tsbad-vuspr}
\end{figure}

Experiments on TSB-AD and TAB test whether retaining predictive evidence improves
task-specific anomaly detection. Controlled ablations examine evidence preservation
and adaptation, while dataset-level analyses and visualizations investigate how
detectors use complementary responses.

\begin{table}[!ht]
\centering
\caption{Four-metric results on the 16 datasets selected by TimeRCD (\%).
\(\diamond\): zero-shot. Red bold and blue underlined italics mark the best
and second-best results per dataset--metric column.}
\label{tab:timercd-full}
\scriptsize
\definecolor{rankFirst}{RGB}{166,38,49}
\definecolor{rankSecond}{RGB}{35,83,151}
\newcommand{\rankbest}[1]{\textcolor{rankFirst}{\textbf{#1}}}
\newcommand{\ranksecond}[1]{\textcolor{rankSecond}{\underline{\textit{#1}}}}
\setlength{\tabcolsep}{1.0pt}
\renewcommand{\arraystretch}{0.7803}
\resizebox{\textwidth}{!}{%
\begin{tabular}{ll*{11}{c}|*{5}{c}|*{2}{>{\centering\arraybackslash}p{2.5em}}}
\toprule
& & \multicolumn{11}{c}{Univariate datasets} & \multicolumn{5}{c}{Multivariate datasets} & \multicolumn{2}{c}{Rank counts} \\
\cmidrule(lr){3-13}\cmidrule(lr){14-18}\cmidrule(lr){19-20}
Metric & Model & IOPS & MGAB & NAB & NEK & Power & SED & Stock & TODS & UCR & WSD & YAHOO & MSL & PSM & SMAP & SMD & SWaT & \textcolor{rankFirst}{1st} & \textcolor{rankSecond}{2nd} \\
\midrule
\rowcolor{black!6}
\multirow{11}{*}{Affiliation-F}
 & \textbf{HieraNTP} & \rankbest{93.68} & 67.67 & \rankbest{96.53} & \rankbest{95.25} & 68.39 & \rankbest{96.20} & \rankbest{97.86} & 87.57 & \rankbest{86.13} & \rankbest{97.96} & 95.19 & 83.37 & \ranksecond{77.87} & \ranksecond{89.91} & 84.47 & 73.41 & \textcolor{rankFirst}{07} & \textcolor{rankSecond}{02} \\
 & TimeRCD~\(\diamond\) & 86.98 & \rankbest{74.54} & \ranksecond{93.81} & 83.95 & \rankbest{85.02} & \ranksecond{88.99} & 41.31 & 84.86 & \ranksecond{84.34} & 91.00 & \ranksecond{95.94} & \ranksecond{84.66} & \rankbest{82.31} & 86.01 & \ranksecond{91.75} & 69.43 & \textcolor{rankFirst}{03} & \textcolor{rankSecond}{06} \\
 & MOMENT~\(\diamond\) & 87.54 & 66.76 & 90.45 & 92.26 & 75.97 & 59.13 & 45.26 & 59.76 & 75.77 & \ranksecond{95.39} & 79.99 & 74.55 & 65.79 & 77.42 & 74.00 & 70.17 & \textcolor{rankFirst}{00} & \textcolor{rankSecond}{01} \\
 & TimesFM~\(\diamond\) & 81.88 & 66.95 & 79.73 & 90.49 & 69.88 & 67.14 & \ranksecond{97.53} & \ranksecond{89.08} & 70.03 & 78.97 & 91.28 & 20.35 & 71.24 & 45.44 & 62.85 & 44.37 & \textcolor{rankFirst}{00} & \textcolor{rankSecond}{02} \\
 & Chronos~\(\diamond\) & \ranksecond{90.12} & 67.89 & 86.66 & \ranksecond{93.63} & 69.72 & 67.89 & 96.85 & \rankbest{91.96} & 74.35 & 90.98 & \rankbest{96.34} & 75.52 & 70.88 & 72.22 & 75.31 & 70.43 & \textcolor{rankFirst}{02} & \textcolor{rankSecond}{02} \\
 & MovingVar.~\(\diamond\) & 83.90 & 67.56 & 91.50 & 77.81 & 74.77 & 67.14 & 67.96 & 71.57 & 78.44 & 89.13 & 78.00 & \rankbest{89.43} & 73.91 & 88.14 & 81.33 & \ranksecond{75.55} & \textcolor{rankFirst}{01} & \textcolor{rankSecond}{01} \\
 & TranAD & 83.19 & 67.28 & 90.28 & 85.02 & 71.56 & 61.03 & 57.94 & 52.76 & 73.31 & 84.34 & 76.08 & 79.91 & 73.83 & 87.39 & \rankbest{92.20} & 75.37 & \textcolor{rankFirst}{01} & \textcolor{rankSecond}{00} \\
 & USAD & 71.08 & 67.81 & 91.54 & 71.13 & 76.48 & 55.60 & 35.92 & 47.90 & 76.00 & 65.10 & 53.05 & 81.86 & 57.86 & 87.25 & 85.09 & 75.06 & \textcolor{rankFirst}{00} & \textcolor{rankSecond}{00} \\
 & OmniAnomaly & 80.32 & 67.35 & 92.35 & 86.30 & \ranksecond{78.16} & 61.26 & 75.24 & 50.73 & 73.53 & 78.02 & 71.31 & 83.15 & 58.17 & \rankbest{91.38} & 85.82 & 73.39 & \textcolor{rankFirst}{01} & \textcolor{rankSecond}{01} \\
 & DCdetector & 71.83 & \ranksecond{67.91} & 72.21 & 62.31 & 69.75 & 72.20 & 55.79 & 57.81 & 70.18 & 72.79 & 67.77 & 67.74 & 67.32 & 67.10 & 69.55 & 71.07 & \textcolor{rankFirst}{00} & \textcolor{rankSecond}{01} \\
 & TFMAE & 78.25 & 67.50 & 75.99 & 76.91 & 70.30 & 68.17 & 56.39 & 62.83 & 70.60 & 80.25 & 76.87 & 75.70 & 70.07 & 75.36 & 70.85 & \rankbest{75.72} & \textcolor{rankFirst}{01} & \textcolor{rankSecond}{00} \\
\cmidrule(lr){1-20}
\rowcolor{black!6}
\multirow{11}{*}{F1-T}
 & \textbf{HieraNTP} & \rankbest{53.84} & 2.30 & \rankbest{72.34} & 75.51 & 19.43 & \rankbest{65.33} & \rankbest{95.62} & \ranksecond{64.24} & \ranksecond{37.77} & \rankbest{64.88} & 78.04 & \rankbest{49.87} & 27.19 & \rankbest{54.21} & 44.66 & 46.04 & \textcolor{rankFirst}{07} & \textcolor{rankSecond}{02} \\
 & TimeRCD~\(\diamond\) & 45.39 & \rankbest{9.05} & 50.49 & \ranksecond{76.91} & 23.05 & \ranksecond{60.59} & 18.38 & \rankbest{68.46} & \rankbest{41.18} & \ranksecond{44.13} & \ranksecond{79.14} & 43.67 & \rankbest{43.86} & 35.82 & \rankbest{55.96} & 35.47 & \textcolor{rankFirst}{05} & \textcolor{rankSecond}{04} \\
 & MOMENT~\(\diamond\) & 33.15 & 0.80 & 52.27 & 63.66 & 19.91 & 9.54 & 18.04 & 17.47 & 13.02 & 41.98 & 11.69 & 25.97 & 27.77 & 17.93 & 28.68 & 28.76 & \textcolor{rankFirst}{00} & \textcolor{rankSecond}{00} \\
 & TimesFM~\(\diamond\) & 48.95 & 0.93 & 36.74 & 36.63 & 19.80 & 9.58 & 88.94 & 51.13 & 10.78 & 41.38 & \rankbest{83.46} & 7.83 & 25.42 & 11.64 & 18.65 & 21.39 & \textcolor{rankFirst}{01} & \textcolor{rankSecond}{00} \\
 & Chronos~\(\diamond\) & 45.45 & 1.10 & 36.10 & 33.16 & 19.90 & 13.18 & \ranksecond{89.30} & 53.90 & 10.88 & 39.82 & 79.00 & 15.59 & 25.42 & 11.72 & 17.32 & 28.88 & \textcolor{rankFirst}{00} & \textcolor{rankSecond}{01} \\
 & MovingVar.~\(\diamond\) & 22.53 & 0.85 & 56.66 & 55.72 & 19.93 & 9.59 & 17.53 & 16.43 & 12.85 & 22.01 & 13.89 & 41.08 & 26.13 & 24.13 & 32.67 & 27.11 & \textcolor{rankFirst}{00} & \textcolor{rankSecond}{00} \\
 & TranAD & 22.63 & 1.65 & 37.28 & 69.97 & 22.36 & 9.57 & 16.73 & 13.51 & 7.75 & 20.94 & 8.41 & 39.42 & 25.49 & 29.12 & 37.98 & \ranksecond{49.58} & \textcolor{rankFirst}{00} & \textcolor{rankSecond}{01} \\
 & USAD & 20.99 & \ranksecond{4.07} & \ranksecond{61.46} & 70.64 & \rankbest{28.23} & 9.54 & 16.86 & 20.85 & 14.63 & 14.18 & 9.35 & 48.71 & 28.96 & 43.94 & 50.41 & \rankbest{50.41} & \textcolor{rankFirst}{02} & \textcolor{rankSecond}{02} \\
 & OmniAnomaly & \ranksecond{51.17} & 1.61 & 40.09 & \rankbest{82.20} & \ranksecond{23.48} & 9.68 & 36.22 & 14.33 & 8.47 & 34.79 & 24.16 & \ranksecond{49.36} & \ranksecond{30.42} & \ranksecond{46.63} & \ranksecond{51.84} & 46.64 & \textcolor{rankFirst}{01} & \textcolor{rankSecond}{06} \\
 & DCdetector & 6.61 & 1.32 & 32.72 & 29.21 & 21.13 & 10.53 & 16.07 & 16.40 & 6.62 & 7.32 & 6.81 & 23.24 & 25.34 & 15.73 & 9.47 & 28.64 & \textcolor{rankFirst}{00} & \textcolor{rankSecond}{00} \\
 & TFMAE & 19.41 & 1.07 & 33.04 & 31.11 & 20.18 & 11.82 & 22.15 & 16.48 & 5.90 & 19.11 & 23.85 & 25.28 & 25.36 & 19.39 & 10.13 & 28.46 & \textcolor{rankFirst}{00} & \textcolor{rankSecond}{00} \\
\cmidrule(lr){1-20}
\rowcolor{black!6}
\multirow{11}{*}{Standard-F1}
 & \textbf{HieraNTP} & \rankbest{54.26} & 2.00 & \rankbest{71.37} & \rankbest{81.49} & 19.40 & \rankbest{65.59} & \rankbest{95.60} & \ranksecond{61.25} & \rankbest{35.47} & \rankbest{64.95} & \ranksecond{85.17} & 37.04 & 26.87 & \rankbest{51.34} & 52.54 & \rankbest{64.02} & \textcolor{rankFirst}{09} & \textcolor{rankSecond}{02} \\
 & TimeRCD~\(\diamond\) & \ranksecond{47.28} & \rankbest{8.63} & 42.66 & 68.64 & 23.12 & \ranksecond{60.85} & 19.56 & \rankbest{70.02} & \ranksecond{33.47} & 41.04 & \rankbest{89.05} & 34.48 & 28.25 & 25.36 & \ranksecond{53.82} & 34.44 & \textcolor{rankFirst}{03} & \textcolor{rankSecond}{04} \\
 & MOMENT~\(\diamond\) & 30.69 & 0.67 & 44.75 & 63.85 & 16.39 & 3.36 & 19.38 & 14.64 & 9.00 & \ranksecond{41.42} & 10.54 & 14.43 & 23.83 & 12.92 & 29.78 & 21.30 & \textcolor{rankFirst}{00} & \textcolor{rankSecond}{01} \\
 & TimesFM~\(\diamond\) & 34.28 & 0.83 & 26.46 & 38.15 & 16.73 & 2.96 & 89.13 & 40.08 & 7.86 & 38.50 & 84.44 & 5.75 & 22.18 & 10.46 & 18.65 & 22.84 & \textcolor{rankFirst}{00} & \textcolor{rankSecond}{00} \\
 & Chronos~\(\diamond\) & 32.69 & 0.99 & 26.22 & 33.54 & 17.47 & 8.74 & \ranksecond{89.41} & 40.52 & 8.21 & 34.58 & 78.89 & 11.63 & 22.27 & 9.62 & 17.50 & 24.03 & \textcolor{rankFirst}{00} & \textcolor{rankSecond}{01} \\
 & MovingVar.~\(\diamond\) & 22.74 & 0.74 & 51.37 & 49.19 & 19.94 & 9.60 & 19.01 & 16.10 & 9.51 & 22.57 & 11.62 & 35.43 & 26.10 & 23.66 & 38.43 & 26.86 & \textcolor{rankFirst}{00} & \textcolor{rankSecond}{00} \\
 & TranAD & 34.85 & 1.46 & 27.33 & 60.36 & 22.36 & 2.63 & 16.23 & 11.94 & 4.40 & 20.23 & 5.70 & 29.60 & 25.63 & 25.11 & 43.99 & 61.86 & \textcolor{rankFirst}{00} & \textcolor{rankSecond}{00} \\
 & USAD & 30.66 & \ranksecond{3.89} & \ranksecond{56.15} & 62.91 & \rankbest{28.24} & 3.41 & 17.99 & 23.87 & 10.74 & 13.20 & 7.21 & \ranksecond{38.71} & \ranksecond{28.41} & 38.66 & 53.06 & \ranksecond{62.82} & \textcolor{rankFirst}{01} & \textcolor{rankSecond}{05} \\
 & OmniAnomaly & 47.05 & 1.44 & 28.81 & \ranksecond{74.03} & \ranksecond{23.50} & 0.43 & 38.59 & 12.65 & 5.11 & 29.57 & 21.40 & \rankbest{39.10} & \rankbest{30.43} & \ranksecond{40.50} & \rankbest{57.06} & 55.93 & \textcolor{rankFirst}{03} & \textcolor{rankSecond}{03} \\
 & DCdetector & 5.19 & 1.21 & 24.02 & 17.37 & 21.10 & 10.54 & 16.97 & 17.85 & 3.18 & 4.64 & 4.14 & 14.08 & 25.33 & 10.67 & 8.99 & 27.02 & \textcolor{rankFirst}{00} & \textcolor{rankSecond}{00} \\
 & TFMAE & 9.48 & 0.97 & 23.78 & 19.74 & 20.14 & 11.87 & 23.32 & 14.88 & 2.83 & 15.53 & 20.50 & 15.68 & 25.39 & 12.58 & 9.16 & 27.08 & \textcolor{rankFirst}{00} & \textcolor{rankSecond}{00} \\
\cmidrule(lr){1-20}
\rowcolor{black!6}
\multirow{11}{*}{VUS-PR}
 & \textbf{HieraNTP} & \rankbest{51.53} & \ranksecond{0.96} & \rankbest{75.16} & \rankbest{85.30} & 10.97 & \rankbest{84.89} & \ranksecond{98.18} & \ranksecond{81.50} & \rankbest{34.92} & \rankbest{64.32} & 79.04 & \rankbest{33.97} & 17.42 & \rankbest{52.37} & \ranksecond{39.22} & 43.17 & \textcolor{rankFirst}{08} & \textcolor{rankSecond}{04} \\
 & TimeRCD~\(\diamond\) & \ranksecond{42.98} & \rankbest{3.09} & 44.65 & 71.13 & \ranksecond{14.63} & \ranksecond{69.86} & 72.73 & \rankbest{91.67} & \ranksecond{27.48} & 29.95 & \rankbest{87.88} & 23.41 & \rankbest{21.69} & 19.68 & \rankbest{46.70} & 18.47 & \textcolor{rankFirst}{05} & \textcolor{rankSecond}{04} \\
 & MOMENT~\(\diamond\) & 37.35 & 0.56 & 45.38 & 67.74 & 10.50 & 4.31 & 76.97 & 56.45 & 6.17 & \ranksecond{55.26} & 30.81 & 9.32 & 16.48 & 8.97 & 15.96 & 14.90 & \textcolor{rankFirst}{00} & \textcolor{rankSecond}{01} \\
 & TimesFM~\(\diamond\) & 19.56 & 0.58 & 24.01 & 35.02 & 10.44 & 6.13 & \rankbest{98.39} & 72.89 & 6.03 & 21.57 & \ranksecond{86.78} & 11.84 & 14.76 & 16.95 & 13.02 & 19.43 & \textcolor{rankFirst}{01} & \textcolor{rankSecond}{01} \\
 & Chronos~\(\diamond\) & 19.00 & 0.60 & 23.76 & 31.80 & 10.95 & 8.65 & 97.49 & 70.66 & 6.56 & 18.81 & 83.54 & 8.25 & 14.61 & 5.18 & 10.22 & 16.44 & \textcolor{rankFirst}{00} & \textcolor{rankSecond}{00} \\
 & MovingVar.~\(\diamond\) & 22.47 & 0.44 & 48.75 & 38.81 & 10.70 & 5.17 & 73.99 & 51.58 & 6.90 & 22.83 & 19.85 & 25.02 & 16.45 & 12.73 & 21.43 & 14.49 & \textcolor{rankFirst}{00} & \textcolor{rankSecond}{00} \\
 & TranAD & 21.61 & 0.64 & 24.82 & 61.63 & 13.04 & 5.75 & 78.08 & 47.33 & 2.25 & 12.20 & 25.78 & 14.78 & 16.49 & 13.37 & 28.34 & \rankbest{47.37} & \textcolor{rankFirst}{01} & \textcolor{rankSecond}{00} \\
 & USAD & 16.58 & 0.75 & \ranksecond{55.03} & 58.53 & \rankbest{18.68} & 4.37 & 74.53 & 56.36 & 8.85 & 10.00 & 14.15 & 29.95 & 17.59 & 26.37 & 34.53 & \ranksecond{44.73} & \textcolor{rankFirst}{01} & \textcolor{rankSecond}{02} \\
 & OmniAnomaly & 25.35 & 0.64 & 27.17 & \ranksecond{74.51} & 14.32 & 6.20 & 91.29 & 45.55 & 2.40 & 16.37 & 29.26 & \ranksecond{31.57} & \ranksecond{18.58} & \ranksecond{28.07} & 37.44 & 42.97 & \textcolor{rankFirst}{00} & \textcolor{rankSecond}{04} \\
 & DCdetector & 5.83 & 0.59 & 16.60 & 14.03 & 12.32 & 9.37 & 74.16 & 46.66 & 1.53 & 3.23 & 10.17 & 7.01 & 14.49 & 4.21 & 4.66 & 15.04 & \textcolor{rankFirst}{00} & \textcolor{rankSecond}{00} \\
 & TFMAE & 5.32 & 0.64 & 15.68 & 17.81 & 11.90 & 9.55 & 73.54 & 48.79 & 2.57 & 5.36 & 25.93 & 8.25 & 14.22 & 5.76 & 4.77 & 15.38 & \textcolor{rankFirst}{00} & \textcolor{rankSecond}{00} \\
\bottomrule
\end{tabular}%
}
\end{table}

\subsection{Experimental Setup}

\paragraph{Benchmarks and supervision.}
We evaluate HieraNTP on TSB-AD \citep{liu2024tsbad} and TAB
\citep{qiu2025tab}, covering univariate detection (TSB-AD-U and TAB-U) and
multivariate detection (TSB-AD-M and TAB-M). TSB-AD provides 48 Tuning and
350 Eval series for TSB-AD-U, and 20 Tuning and 180 Eval series for TSB-AD-M.
TAB provides official training/test splits for 1635 univariate series from
15 datasets and 220 multivariate series from 29 datasets.

Normal and anomaly labels from the TSB-AD Tuning sets and TAB training data
supervise the shared base detector and dataset-specific extensions.
TSB-AD uses sequence-level cross-validation within Tuning to select
hyperparameters and extensions, with temporal holdouts for eligible singleton
datasets. TAB follows its full-shot evaluation strategy with fixed
hyperparameters shared across datasets. HieraNTP pretrains on QUALS, derived
from GIFT-Eval Pretrain \citep{li2027quals,aksu2024gifteval}, with no data
leakage from TSB-AD or TAB. The tokenizer and predictor remain frozen.

% Source: TAB25, opents_base384_family32_recomputed_20260924.
% Public 14/28-dataset scope; OURS_FAMILY32 inserted alone (384 + family 32).
\begin{table}[t]
\centering
\caption{TAB full-shot results. Red bold: best; blue underlined italics: second best.}
\label{tab:opents-main}
% Match Table 1's rendered size: 7pt times its 0.92471 resize factor.
\fontsize{6.473}{8}\selectfont
\definecolor{tabFirst}{RGB}{166,38,49}
\definecolor{tabSecond}{RGB}{35,83,151}
\newcommand{\tabbest}[1]{\textcolor{tabFirst}{\textbf{#1}}}
\newcommand{\tabsecond}[1]{\textcolor{tabSecond}{\underline{\textit{#1}}}}
\setlength{\tabcolsep}{1pt}
\renewcommand{\arraystretch}{0.697}
\newcommand{\tabours}{HieraNTP}
\newcommand{\tabmodel}[1]{#1}
\begin{tabularx}{\textwidth}{@{}l>{\columncolor{black!6}\centering\arraybackslash}X*{11}{>{\centering\arraybackslash}X}@{}}
\toprule
\multicolumn{13}{c}{\textbf{Univariate Detection with 14 datasets}} \\
\midrule
Metric
& \tabours
& \tabmodel{KMeans}
& \tabmodel{\shortstack[c]{DWT-\\MLEAD}}
& \tabmodel{CALF}
& \tabmodel{\shortstack[c]{iTrans\\former}}
& \tabmodel{EIF}
& \tabmodel{ContraAD}
& \tabmodel{DLinear}
& \tabmodel{TranAD}
& \tabmodel{\shortstack[c]{DC\\detector}}
& \tabmodel{DualTF}
& \tabmodel{\shortstack[c]{Modern\\TCN}} \\
\midrule
F1 & \tabbest{0.322} & \tabsecond{0.293} & 0.242 & 0.182 & 0.194 & 0.166 & 0.041 & 0.165 & 0.158 & 0.024 & 0.078 & 0.163 \\
Aff-F1 & \tabbest{0.849} & 0.792 & 0.716 & \tabsecond{0.820} & 0.810 & 0.782 & 0.684 & 0.808 & 0.774 & 0.688 & 0.728 & 0.807 \\
AP & \tabbest{0.419} & \tabsecond{0.356} & 0.291 & 0.231 & 0.243 & 0.213 & 0.033 & 0.213 & 0.201 & 0.034 & 0.083 & 0.210 \\
AUC-ROC & \tabbest{0.817} & \tabsecond{0.790} & 0.739 & 0.687 & 0.687 & 0.633 & 0.499 & 0.659 & 0.545 & 0.473 & 0.539 & 0.663 \\
VUS-PR & \tabbest{0.439} & \tabsecond{0.362} & 0.312 & 0.251 & 0.244 & 0.202 & 0.059 & 0.202 & 0.179 & 0.057 & 0.108 & 0.210 \\
VUS-ROC & \tabbest{0.837} & \tabsecond{0.784} & 0.699 & 0.758 & 0.753 & 0.671 & 0.579 & 0.721 & 0.594 & 0.491 & 0.604 & 0.725 \\
$G/S/B$ & 62/18/29 & 28/62/27 & 29/22/22 & 8/14/7 & 3/2/24 & 4/4/8 & 1/8/3 & 1/3/5 & 0/2/6 & 1/2/0 & 0/1/3 & 1/0/0 \\
Score & \tabbest{251} & \tabsecond{235} & 153 & 59 & 37 & 28 & 22 & 14 & 10 & 7 & 5 & 3 \\
\bottomrule
\end{tabularx}

\begin{tabularx}{\textwidth}{@{}l>{\columncolor{black!6}\centering\arraybackslash}X*{11}{>{\centering\arraybackslash}X}@{}}
\toprule
\multicolumn{13}{c}{\textbf{Multivariate Detection with 28 datasets}} \\
\midrule
Metric
& \tabours
& \tabmodel{KMeans}
& \tabmodel{KNN}
& \tabmodel{EIF}
& \tabmodel{OCSVM}
& \tabmodel{TimesNet}
& \tabmodel{CATCH}
& \tabmodel{\shortstack[c]{iTrans\\former}}
& \tabmodel{DLinear}
& \tabmodel{DUET}
& \tabmodel{DualTF}
& \tabmodel{\shortstack[c]{DC\\detector}} \\
\midrule
F1 & \tabbest{0.331} & \tabsecond{0.296} & 0.179 & 0.210 & 0.161 & 0.199 & 0.191 & 0.183 & 0.176 & 0.200 & 0.109 & 0.046 \\
Aff-F1 & \tabbest{0.793} & 0.698 & 0.699 & 0.773 & 0.677 & \tabsecond{0.789} & 0.778 & 0.770 & 0.775 & 0.780 & 0.593 & 0.637 \\
AP & \tabbest{0.376} & \tabsecond{0.340} & 0.336 & 0.254 & 0.176 & 0.259 & 0.240 & 0.235 & 0.227 & 0.239 & 0.147 & 0.085 \\
AUC-ROC & \tabbest{0.782} & \tabsecond{0.744} & 0.714 & 0.732 & 0.645 & 0.688 & 0.668 & 0.667 & 0.660 & 0.663 & 0.611 & 0.496 \\
VUS-PR & \tabbest{0.388} & 0.302 & 0.301 & 0.258 & \tabsecond{0.309} & 0.271 & 0.261 & 0.251 & 0.240 & 0.258 & 0.174 & 0.118 \\
VUS-ROC & \tabbest{0.794} & 0.685 & 0.707 & \tabsecond{0.741} & 0.656 & 0.722 & 0.705 & 0.705 & 0.695 & 0.699 & 0.621 & 0.503 \\
$G/S/B$ & 108/46/25 & 56/64/25 & 60/21/35 & 27/40/25 & 41/19/15 & 26/17/40 & 10/15/27 & 7/16/24 & 9/14/10 & 4/6/22 & 7/4/10 & 2/4/7 \\
Score & \tabbest{441} & \tabsecond{321} & 257 & 186 & 176 & 152 & 87 & 77 & 65 & 46 & 39 & 21 \\
\bottomrule
\end{tabularx}
\end{table}

\paragraph{Comparisons.}
TSB-AD comparisons follow the official Tuning/Eval split.
Figure~\ref{fig:tsbad-vuspr} compares leading leaderboard methods, and
Table~\ref{tab:timercd-full} reports results on the 16 datasets selected by
TimeRCD \citep{liu2024tsbad,lan2025timercd}.
Table~\ref{tab:opents-main} compares TAB results on the 14 univariate and 28
multivariate datasets in OpenTS, excluding OPP. These baselines combine
leading leaderboard methods with deep-learning detectors. Classical baselines
include KMeans, KNN, EIF, OCSVM, and DWT-MLEAD
\citep{macqueen1967some,ramaswamy2000efficient,hariri2021extended,scholkopf1999support,thill2017time}.
Neural comparisons cover DLinear, iTransformer, TimesNet, ModernTCN, DUET,
and CALF \citep{zeng2023transformers,liu2024itransformer,wu2023timesnet,luo2024moderntcn,qiu2025duet,liu2025calf},
alongside TranAD, DCdetector, DualTF, ContraAD, and CATCH
\citep{tuli2022tranad,yang2023dcdetector,nam2024breaking,zhuang2024noise,wu2025catch}.

\paragraph{Metrics and implementation.}
VUS-PR is the primary metric \citep{paparrizos2022vus}, supplemented by
AUC-PR (average precision, AP), VUS-ROC, AUC-ROC, and F1 metrics including
Affiliation-F \citep{huet2022affiliation}. For TAB, the composite score is $3G+2S+B$:
$G$, $S$, and $B$ count first-, second-, and third-place finishes across
dataset--metric pairs for all 16 metrics, with ties sharing gold, silver,
or bronze medals. Appendices~\ref{app:implementation}--\ref{app:additional-ablations}
detail training, evaluation, and results.

\begin{table}[!t]
\centering
\caption{Evidence and adaptation analysis on TSB-AD-U/M (VUS-PR, \%).}
\label{tab:component-ablation}
\scriptsize
\renewcommand{\arraystretch}{0.84}
\setlength{\tabcolsep}{1pt}
\renewcommand{\tabularxcolumn}[1]{m{#1}}
\begin{tabularx}{\textwidth}{@{}
>{\raggedright\arraybackslash}X
*{2}{>{\centering\arraybackslash}m{0.047\textwidth}}
@{\hspace{5pt}}
>{\raggedright\arraybackslash}X
*{2}{>{\centering\arraybackslash}m{0.047\textwidth}}
@{\hspace{5pt}}
>{\raggedright\arraybackslash}X
*{2}{>{\centering\arraybackslash}m{0.047\textwidth}}@{}}
\toprule
\multicolumn{3}{c}{Evidence construction} &
\multicolumn{3}{c}{Temporal structure} &
\multicolumn{3}{c}{Scoring and adaptation} \\
\cmidrule(lr){1-3}\cmidrule(lr){4-6}\cmidrule(l){7-9}
Configuration & U & M & Configuration & U & M & Configuration & U & M \\
\midrule
\textbf{HieraNTP} & 65.56 & 50.28 &
\textbf{HieraNTP} & 65.56 & 50.28 &
\textbf{HieraNTP} & 65.56 & 50.28 \\
Prediction-to-observation only & 59.78 & 44.60 & w/o patch 256 & 65.53 & 49.98 &
Mean + detector & 38.20 & 29.12 \\
w/o token-space & 65.35 & 49.69 & w/o patch 128 & 65.52 & 49.29 &
Max + detector & 32.11 & 27.47 \\
w/o patch mean & 64.31 & 50.33 & w/o patch 64 & 65.83 & 49.92 &
Mean--max + detector & 44.51 & 33.83 \\
w/o patch variance & 62.47 & 47.25 & w/o patch 32 & 66.36 & 50.08 &
Time-domain error & 33.85 & 30.81 \\
w/o observation-space & 62.54 & 47.77 & w/o patch 16 & 66.12 & 49.70 & Time-domain error + detector & 29.89 & 28.34 \\
w/o reference calibration & 60.28 & 51.71 & w/o patch 8 & 63.46 & 49.46 & Instantaneous multiview & 57.80 & 44.08 \\
Variable mean only & -- & 47.55 & Single token-grid offset & 64.22 & 49.62 & Fixed evidence averaging & 43.65 & 35.27 \\
Variable maximum only & -- & 47.41 & w/o temporal summaries & 59.25 & 45.63 & Shared base detector only & 61.48 & 49.91 \\
& & & Forward only & 60.32 & 47.52 & & & \\
\bottomrule
\end{tabularx}
\end{table}

\begin{figure}[t]
  \centering
  \includegraphics[width=\linewidth]{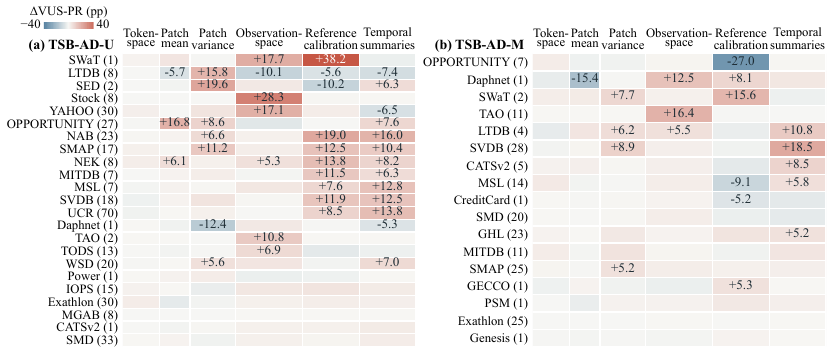}
  \caption{Dataset-wise VUS-PR differences: HieraNTP minus component removal
  (pp). Numbers mark magnitudes $\geq5$; parentheses give sequence counts.
  Rows are clustered within each track.}
  \label{fig:dataset-ablation}
\end{figure}

\subsection{Main Results}
\label{sec:tsbad-main-results}
\label{sec:opents-results}

Figure~\ref{fig:tsbad-vuspr} shows that HieraNTP achieves the highest VUS-PR
on both official TSB-AD Eval sets, improving on the strongest baselines
by \HieraGainU{} on U and \HieraGainM{} on M in relative terms.
Across the 16 datasets in Table~\ref{tab:timercd-full}, it ranks first in
VUS-PR on eight and second on four, and obtains the most first-place results
for each F1 metric. Gains vary by dataset: on SMAP, VUS-PR exceeds the
strongest baseline by 24.30 percentage points, while TimeRCD leads on TODS and SMD.
Joint gains in ranking and F1 metrics indicate better anomaly ordering and
threshold-based detection.

On TAB, HieraNTP leads on all six metrics in both tracks
in Table~\ref{tab:opents-main}. Dataset-mean
VUS-PR exceeds the strongest baseline by 7.7 percentage points on univariate
detection and 7.9 on multivariate detection. HieraNTP also ranks first in
composite score, with 251 on U and 441 on M. Together, the two benchmarks
support learning anomaly decisions from structured predictive evidence across
different datasets and evaluation protocols, with the pretrained temporal model
frozen.

\subsection{Ablation Study}
\label{sec:risk-analysis}

Table~\ref{tab:component-ablation} examines evidence content, temporal organization,
and adaptation with pretrained modules frozen. Supervised comparisons fix
samples, labels, weights, and detector hyperparameters. Time-domain error is
scored directly or supplied to the detector. Fixed evidence averaging takes
each Pool$\to$Cal. channel's positive part before temporal summaries, then
averages channels and directions without fitting a detector.

\paragraph{Evidence content and compression.}
Under matched supervised training, the complete representation outperforms
mean, maximum, and mean--max inputs, improving VUS-PR over mean--max by
21.05 percentage points on U and 16.45 on M. Instantaneous multiview evidence retains
64 current-value features on U and 128 on M after variable mean/max aggregation.
It also outperforms the time-domain error input. These results support retaining
distinct predictive responses for task-specific decision learning to address
the predictive risk--decision gap.

Prediction-to-observation only retains decoded-observation evidence at native
scales and full resolution with unchanged processing. Restoring the remaining
sources improves VUS-PR by 5.78
percentage points on U and 5.68 on M. Token-space removal has smaller effects
and retains local-statistic and observation-space predictions. These comparisons
support complementary views of the same predictive process.
On M, combining variable means and maxima outperforms either alone, supporting
both broad and concentrated responses across variables.

\begin{figure}[t]
  \centering
  \includegraphics[width=\linewidth]{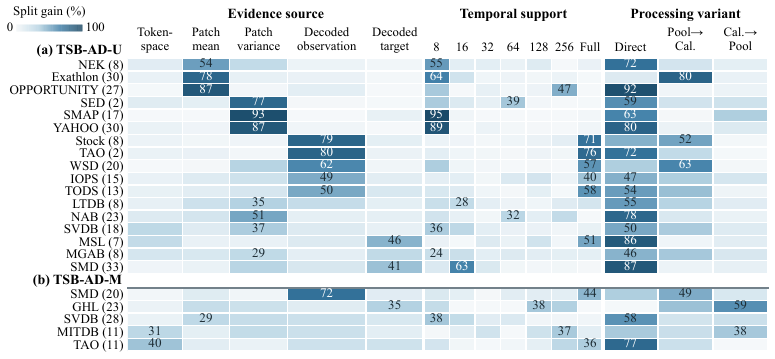}
  \caption{Evidence usage in dataset-specific extensions. Split-gain shares
  (\%) are normalized by column group; numbers mark row-wise maxima per group.
  Full denotes full resolution; parentheses give evaluation sequence counts.
  Rows are clustered by source profiles within each track.}
  \label{fig:dataset-evidence-usage}
\end{figure}

\begin{figure}[t]
  \centering
  \includegraphics[width=\linewidth]{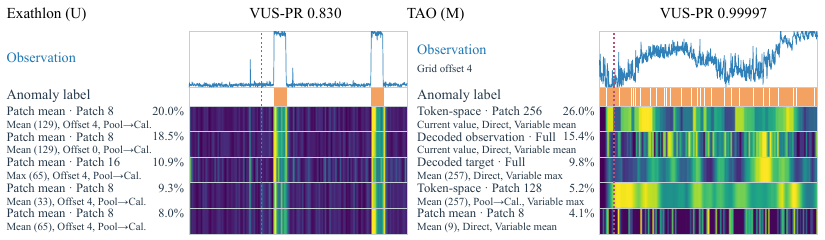}
  \caption{Top-five features in Exathlon (U) and TAO (M). Colors are scaled per
  feature; percentages show dataset-specific split gain. VUS-PR is computed
  per window.}
  \label{fig:selected-evidence}
\end{figure}

\paragraph{Temporal organization and calibration.}
Retaining only current values from all three processing paths reduces VUS-PR
by 6.31 percentage points on U and 4.64 on M. Temporal means and maxima
preserve sustained and peak responses. Their benefit supports evaluating how
predictive responses persist and concentrate over time when learning which
deviations satisfy the target anomaly criterion. Paired sequence-resampling
intervals in Table~\ref{tab:paired-ablation}
remain above zero for this removal and for patch-variance and observation-space
removal. Removing reference calibration lowers U by 5.28 points but raises M
by 1.43 points.

Patch-evidence removals retain the pretrained hierarchy. Removing patch 8
lowers U by 2.10 percentage points, while removing patch 32 raises it by 0.80
points. Patch-128 removal causes the largest patch-wise decline on M.
A single token-grid offset or forward-direction evidence alone also lowers
both track averages. Temporal views provide unequal benefits across tracks.

\paragraph{Dataset-specific adaptation.}
With the complete representation fixed, dataset-specific extensions improve
VUS-PR over the shared base detector by 4.08 percentage points on U and 0.37
on M. Since the predictive responses are unchanged, the improvement comes from
refining their combination for the target dataset. This supports retaining
shared evidence while learning which combinations correspond to each dataset's
anomaly criteria.

\subsection{Evidence Usage and Visualization}
\label{sec:evidence-usage}

Figure~\ref{fig:dataset-evidence-usage} summarizes extension split gain for
17 univariate and five multivariate datasets. Patch mean dominates Exathlon,
patch variance SMAP, and decoded-observation evidence Stock.
Figure~\ref{fig:dataset-ablation} shows pronounced losses from patch-variance
removal in SMAP and observation-space removal in Stock. Together, these results
support dataset-dependent use of the shared evidence representation.

In Figure~\ref{fig:selected-evidence}, patch-mean evidence tracks the onset and
persistence of two labeled
level shifts in Exathlon. In TAO, token-space responses remain elevated early,
while decoded-observation evidence varies rapidly and temporal averaging smooths
decoded-target responses. These patterns illustrate distinct information
retained for task-adaptive detection, with further cases in
Appendix~\ref{app:additional-cases}.

\section{Conclusion}
\label{sec:conclusion}

HieraNTP addresses the predictive risk--decision gap by separating the modeling
of temporal regularities from the interpretation of their violations. Its
pretrained tokenizer and hierarchical next-token predictor provide multiscale
evidence in token, local-statistic, and observation spaces. Preserving these
discrepancies in a structured risk representation allows a task-adaptive detector
to learn task-specific evidence combinations from anomaly labels while the
pretrained modules remain frozen.

HieraNTP achieves the highest aggregate VUS-PR on both tracks of TSB-AD and
TAB. Under matched detector training, preserving distinct evidence improves
detection over time-domain errors and early aggregation. Ablations and
evidence-use analyses reveal complementary contributions whose value varies
across datasets and temporal configurations. These findings support preserving
predictive evidence until downstream supervision determines its relevance to
the anomaly criterion.

\bibliography{iclr2027_conference}

@inproceedings{liu2021swin,
  title     = {Swin Transformer: Hierarchical Vision Transformer Using Shifted Windows},
  author    = {Liu, Ze and Lin, Yutong and Cao, Yue and Hu, Han and Wei, Yixuan and Zhang, Zheng and Lin, Stephen and Guo, Baining},
  booktitle = {Proceedings of the IEEE/CVF International Conference on Computer Vision},
  year      = {2021}
}

@inproceedings{oord2017neural,
  title     = {Neural Discrete Representation Learning},
  author    = {van den Oord, Aaron and Vinyals, Oriol and Kavukcuoglu, Koray},
  booktitle = {Advances in Neural Information Processing Systems},
  year      = {2017}
}

@article{ledoit2004wellconditioned,
  title   = {A Well-Conditioned Estimator for Large-Dimensional Covariance Matrices},
  author  = {Ledoit, Olivier and Wolf, Michael},
  journal = {Journal of Multivariate Analysis},
  volume  = {88},
  number  = {2},
  pages   = {365--411},
  year    = {2004},
  doi     = {10.1016/S0047-259X(03)00096-4}
}

@inproceedings{chen2016xgboost,
  title     = {{XGBoost}: A Scalable Tree Boosting System},
  author    = {Chen, Tianqi and Guestrin, Carlos},
  booktitle = {Proceedings of the 22nd ACM SIGKDD International Conference on Knowledge Discovery and Data Mining},
  pages     = {785--794},
  year      = {2016},
  doi       = {10.1145/2939672.2939785}
}

@inproceedings{xu2018donut,
  title     = {Unsupervised Anomaly Detection via Variational Auto-Encoder for Seasonal {KPI}s in Web Applications},
  author    = {Xu, Haowen and Chen, Wenxiao and Zhao, Nengwen and Li, Zeyan and Bu, Jiahao and Li, Zhihan and Liu, Ying and Zhao, Youjian and Pei, Dan and Feng, Yang and Chen, Jie and Wang, Zhaogang and Qiao, Honglin},
  booktitle = {Proceedings of The Web Conference},
  year      = {2018},
  doi       = {10.1145/3178876.3185996}
}

@inproceedings{su2019omnianomaly,
  title     = {Robust Anomaly Detection for Multivariate Time Series through Stochastic Recurrent Neural Network},
  author    = {Su, Ya and Zhao, Youjian and Niu, Chenhao and Liu, Rong and Sun, Wei and Pei, Dan},
  booktitle = {Proceedings of the 25th ACM SIGKDD International Conference on Knowledge Discovery and Data Mining},
  year      = {2019},
  doi       = {10.1145/3292500.3330672}
}

@inproceedings{audibert2020usad,
  title     = {{USAD}: UnSupervised Anomaly Detection on Multivariate Time Series},
  author    = {Audibert, Julien and Michiardi, Pietro and Guyard, Fr{\'e}d{\'e}ric and Marti, S{\'e}bastien and Zuluaga, Maria A.},
  booktitle = {Proceedings of the 26th ACM SIGKDD International Conference on Knowledge Discovery and Data Mining},
  year      = {2020},
  doi       = {10.1145/3394486.3403392}
}

@article{tuli2022tranad,
  title   = {{TranAD}: Deep Transformer Networks for Anomaly Detection in Multivariate Time Series Data},
  author  = {Tuli, Shreshth and Casale, Giuliano and Jennings, Nicholas R.},
  journal = {Proceedings of the VLDB Endowment},
  volume  = {15},
  number  = {6},
  pages   = {1201--1214},
  year    = {2022},
  doi     = {10.14778/3514061.3514067}
}

@inproceedings{xu2022anomalytransformer,
  title     = {Anomaly Transformer: Time Series Anomaly Detection with Association Discrepancy},
  author    = {Xu, Jiehui and Wu, Haixu and Wang, Jianmin and Long, Mingsheng},
  booktitle = {International Conference on Learning Representations},
  year      = {2022},
  url       = {https://openreview.net/forum?id=LzQQ89U1qm_}
}

@inproceedings{yang2023dcdetector,
  title     = {{DCdetector}: Dual Attention Contrastive Representation Learning for Time Series Anomaly Detection},
  author    = {Yang, Yiyuan and Zhang, Chaoli and Zhou, Tian and Wen, Qingsong and Sun, Liang},
  booktitle = {Proceedings of the 29th ACM SIGKDD Conference on Knowledge Discovery and Data Mining},
  year      = {2023},
  doi       = {10.1145/3580305.3599295}
}

@inproceedings{yeh2016matrixprofile,
  title     = {Matrix Profile {I}: All Pairs Similarity Joins for Time Series: A Unifying View That Includes Motifs, Discords and Shapelets},
  author    = {Yeh, Chin-Chia Michael and Zhu, Yan and Ulanova, Liudmila and Begum, Nurjahan and Ding, Yifei and Dau, Hoang Anh and Silva, Diego Furtado and Mueen, Abdullah and Keogh, Eamonn J.},
  booktitle = {2016 IEEE 16th International Conference on Data Mining},
  pages     = {1317--1322},
  year      = {2016},
  doi       = {10.1109/ICDM.2016.0179}
}

@inproceedings{fang2024tfmae,
  title     = {Temporal-Frequency Masked Autoencoders for Time Series Anomaly Detection},
  author    = {Fang, Yuchen and Xie, Jiandong and Zhao, Yan and Chen, Lu and Gao, Yunjun and Zheng, Kai},
  booktitle = {2024 IEEE 40th International Conference on Data Engineering},
  pages     = {1228--1241},
  year      = {2024},
  doi       = {10.1109/ICDE60146.2024.00099}
}

@inproceedings{pmlr-v235-goswami24a,
  title     = {{MOMENT}: A Family of Open Time-Series Foundation Models},
  author    = {Goswami, Mononito and Szafer, Konrad and Choudhry, Arjun and Cai, Yifu and Li, Shuo and Dubrawski, Artur},
  booktitle = {Proceedings of the 41st International Conference on Machine Learning},
  series    = {Proceedings of Machine Learning Research},
  volume    = {235},
  year      = {2024},
  url       = {https://proceedings.mlr.press/v235/goswami24a.html}
}

@article{ansari2024chronos,
  title   = {Chronos: Learning the Language of Time Series},
  author  = {Ansari, Abdul Fatir and Stella, Lorenzo and Turkmen, Caner and Zhang, Xiyuan and Mercado, Pedro and Shen, Huibin and Shchur, Oleksandr and Rangapuram, Syama Sundar and Arango, Sebastian Pineda and Kapoor, Shubham and Zschiegner, Jasper and Maddix, Danielle C. and Wang, Hao and Mahoney, Michael W. and Torkkola, Kari and Wilson, Andrew Gordon and Bohlke-Schneider, Michael and Wang, Yuyang},
  journal = {Transactions on Machine Learning Research},
  year    = {2024},
  url     = {https://openreview.net/forum?id=gerNCVqqtR}
}

@inproceedings{pmlr-v235-das24c,
  title     = {A Decoder-Only Foundation Model for Time-Series Forecasting},
  author    = {Das, Abhimanyu and Kong, Weihao and Sen, Rajat and Zhou, Yichen},
  booktitle = {Proceedings of the 41st International Conference on Machine Learning},
  series    = {Proceedings of Machine Learning Research},
  volume    = {235},
  year      = {2024},
  url       = {https://proceedings.mlr.press/v235/das24c.html}
}

@inproceedings{liu2024tsbad,
  title     = {The Elephant in the Room: Towards A Reliable Time-Series Anomaly Detection Benchmark},
  author    = {Liu, Qinghua and Paparrizos, John},
  booktitle = {Advances in Neural Information Processing Systems: Datasets and Benchmarks Track},
  year      = {2024},
  url       = {https://proceedings.neurips.cc/paper_files/paper/2024/hash/c3f3c690b7a99fba16d0efd35cb83b2c-Abstract-Datasets_and_Benchmarks_Track.html}
}

@article{paparrizos2022vus,
  title   = {Volume Under the Surface: A New Accuracy Evaluation Measure for Time-Series Anomaly Detection},
  author  = {Paparrizos, John and Boniol, Paul and Palpanas, Themis and Tsay, Ruey S. and Elmore, Aaron and Franklin, Michael J.},
  journal = {Proceedings of the VLDB Endowment},
  volume  = {15},
  number  = {11},
  pages   = {2774--2787},
  year    = {2022},
  doi     = {10.14778/3551793.3551830},
  url     = {https://doi.org/10.14778/3551793.3551830}
}

@inproceedings{huet2022affiliation,
  title     = {Local Evaluation of Time Series Anomaly Detection Algorithms},
  author    = {Huet, Alexis and Navarro, Jose Manuel and Rossi, Dario},
  booktitle = {Proceedings of the 28th ACM SIGKDD Conference on Knowledge Discovery and Data Mining},
  year      = {2022},
  doi       = {10.1145/3534678.3539339},
  url       = {https://doi.org/10.1145/3534678.3539339}
}

@inproceedings{pmlr-v235-sarfraz24a,
  title     = {Position: Quo Vadis, Unsupervised Time Series Anomaly Detection?},
  author    = {Sarfraz, M. Saquib and Chen, Mei-Yen and Layer, Lukas and Peng, Kunyu and Koulakis, Marios},
  booktitle = {Proceedings of the 41st International Conference on Machine Learning},
  series    = {Proceedings of Machine Learning Research},
  volume    = {235},
  year      = {2024},
  url       = {https://proceedings.mlr.press/v235/sarfraz24a.html}
}

@inproceedings{shentu2025dada,
  title     = {Towards a General Time Series Anomaly Detector with Adaptive Bottlenecks and Dual Adversarial Decoders},
  author    = {Shentu, Qichao and Li, Beibu and Zhao, Kai and Shu, Yang and Rao, Zhongwen and Pan, Lujia and Yang, Bin and Guo, Chenjuan},
  booktitle = {International Conference on Learning Representations},
  year      = {2025},
  url       = {https://openreview.net/forum?id=aKcd7ImG5e}
}

@inproceedings{lan2025timercd,
  title     = {Towards Foundation Models for Zero-Shot Time Series Anomaly Detection: Leveraging Synthetic Data and Relative Context Discrepancy},
  author    = {Lan, Tian and Le, Hao Duong and Li, Jinbo and He, Wenjun and Wang, Meng and Liu, Chenghao and Zhang, Chen},
  booktitle = {Proceedings of the 43rd International Conference on Machine Learning},
  series    = {Proceedings of Machine Learning Research},
  volume    = {306},
  year      = {2026},
  url       = {https://arxiv.org/abs/2509.21190}
}

@inproceedings{ekambaram2025tspulse,
  title     = {{TSPulse}: Tiny Pre-Trained Models with Disentangled Representations for Rapid Time-Series Analysis},
  author    = {Ekambaram, Vijay and Kumar, Subodh and Jati, Arindam and Mukherjee, Sumanta and Sakai, Tomoya and Dayama, Pankaj and Gifford, Wesley M. and Kalagnanam, Jayant},
  booktitle = {International Conference on Learning Representations},
  year      = {2026},
  url       = {https://arxiv.org/abs/2505.13033}
}

@inproceedings{zhu2026oneliners,
  title     = {When Foundation Models Are One-Liners: Limitations and Future Directions for Time Series Anomaly Detection},
  author    = {Zhu, Xiaokun and Carpentier, Louis and Verbeke, Mathias},
  booktitle = {International Conference on Learning Representations},
  year      = {2026},
  url       = {https://proceedings.iclr.cc/paper_files/paper/2026/hash/cf70320e93c08b39b1b29a348097a376-Abstract-Conference.html}
}

@inproceedings{gil2026adaptive,
  title     = {Adaptive Conformal Anomaly Detection with Time Series Foundation Models for Signal Monitoring},
  author    = {Martinez Gil, Natalia and O'Donncha, Fearghal and Gifford, Wesley M. and Zhou, Nianjun and Patel, Dhaval C. and Vaculin, Roman},
  booktitle = {International Conference on Learning Representations},
  year      = {2026},
  url       = {https://openreview.net/forum?id=7uFbs68MSI}
}

@inproceedings{khan2026chronosad,
  title     = {{ChronosAD}: Leveraging Time Series Foundation Models for Accurate Anomaly Detection},
  author    = {Khan, Uzair and Capogrosso, Luigi and Biondani, Francesco and Magno, Michele and Fummi, Franco and Setti, Francesco and Cristani, Marco},
  booktitle = {IEEE International Conference on Industrial Informatics},
  year      = {2026},
  url       = {https://arxiv.org/abs/2606.01300}
}

@article{uray2026exploring,
  title   = {Exploring Zero-Shot Foundation Models for Multivariate Time Series Anomaly Detection},
  author  = {Uray, Martin and Messineo, Saverio and Kwitt, Roland and Huber, Stefan},
  journal = {arXiv preprint arXiv:2607.12454},
  year    = {2026},
  url     = {https://arxiv.org/abs/2607.12454}
}

@article{han2025timerep,
  title   = {Leveraging Intermediate Representations of Time Series Foundation Models for Anomaly Detection},
  author  = {Han, Chan Sik and Lee, Keon Myung},
  journal = {arXiv preprint arXiv:2509.12650},
  year    = {2025},
  url     = {https://arxiv.org/abs/2509.12650}
}

@article{qiu2025tab,
  title   = {{TAB}: Unified Benchmarking of Time Series Anomaly Detection Methods},
  author  = {Qiu, Xiangfei and Li, Zhe and Qiu, Wanghui and Hu, Shiyan and Zhou, Lekui and Wu, Xingjian and Li, Zhengyu and Guo, Chenjuan and Zhou, Aoying and Sheng, Zhenli and Hu, Jilin and Jensen, Christian S. and Yang, Bin},
  journal = {Proceedings of the VLDB Endowment},
  volume  = {18},
  number  = {9},
  pages   = {2775--2789},
  year    = {2025},
  doi     = {10.14778/3746405.3746407},
  url     = {https://www.vldb.org/pvldb/vol18/p2775-hu.pdf}
}

@inproceedings{macqueen1967some,
  title     = {Some Methods for Classification and Analysis of Multivariate Observations},
  author    = {MacQueen, J. B.},
  booktitle = {Proceedings of the Fifth Berkeley Symposium on Mathematical Statistics and Probability},
  volume    = {1},
  pages     = {281--297},
  year      = {1967},
  url       = {https://digicoll.lib.berkeley.edu/record/113015?v=pdf}
}

@inproceedings{thill2017time,
  title     = {Time Series Anomaly Detection with Discrete Wavelet Transforms and Maximum Likelihood Estimation},
  author    = {Thill, Markus and Konen, Wolfgang and B{\"a}ck, Thomas},
  booktitle = {International Work-Conference on Time Series},
  year      = {2017},
  url       = {https://blogs.gm.fh-koeln.de/ciop/files/2019/01/thillwavelet.pdf}
}

@inproceedings{liu2025calf,
  title     = {{CALF}: Aligning {LLM}s for Time Series Forecasting via Cross-modal Fine-Tuning},
  author    = {Liu, Peiyuan and Guo, Hang and Dai, Tao and Li, Naiqi and Bao, Jigang and Ren, Xudong and Jiang, Yong and Xia, Shu-Tao},
  booktitle = {Proceedings of the AAAI Conference on Artificial Intelligence},
  volume    = {39},
  number    = {18},
  pages     = {18915--18923},
  year      = {2025},
  doi       = {10.1609/aaai.v39i18.34082}
}

@inproceedings{liu2024itransformer,
  title     = {{iTransformer}: Inverted Transformers Are Effective for Time Series Forecasting},
  author    = {Liu, Yong and Hu, Tengge and Zhang, Haoran and Wu, Haixu and Wang, Shiyu and Ma, Lintao and Long, Mingsheng},
  booktitle = {International Conference on Learning Representations},
  year      = {2024},
  url       = {https://proceedings.iclr.cc/paper_files/paper/2024/hash/2ea18fdc667e0ef2ad82b2b4d65147ad-Abstract-Conference.html}
}

@article{hariri2021extended,
  title   = {Extended Isolation Forest},
  author  = {Hariri, Sahand and Carrasco Kind, Matias and Brunner, Robert J.},
  journal = {IEEE Transactions on Knowledge and Data Engineering},
  volume  = {33},
  number  = {4},
  pages   = {1479--1489},
  year    = {2021},
  doi     = {10.1109/TKDE.2019.2947676}
}

@article{zhuang2024noise,
  title   = {Noise Matters: Cross Contrastive Learning for Flink Anomaly Detection},
  author  = {Zhuang, Zhihao and Zhang, Yingying and Zhao, Kai and Guo, Chenjuan and Yang, Bin and Wen, Qingsong and Fan, Lunting},
  journal = {Proceedings of the VLDB Endowment},
  volume  = {18},
  number  = {4},
  pages   = {1159--1168},
  year    = {2024},
  doi     = {10.14778/3717755.3717773}
}

@inproceedings{zeng2023transformers,
  title     = {Are Transformers Effective for Time Series Forecasting?},
  author    = {Zeng, Ailing and Chen, Muxi and Zhang, Lei and Xu, Qiang},
  booktitle = {Proceedings of the AAAI Conference on Artificial Intelligence},
  volume    = {37},
  number    = {9},
  pages     = {11121--11128},
  year      = {2023},
  doi       = {10.1609/aaai.v37i9.26317}
}

@inproceedings{nam2024breaking,
  title     = {Breaking the Time-Frequency Granularity Discrepancy in Time-Series Anomaly Detection},
  author    = {Nam, Youngeun and Yoon, Susik and Shin, Yooju and Bae, Minyoung and Song, Hwanjun and Lee, Jae-Gil and Lee, Byung Suk},
  booktitle = {Proceedings of the ACM Web Conference 2024},
  pages     = {4204--4215},
  year      = {2024},
  doi       = {10.1145/3589334.3645556}
}

@inproceedings{luo2024moderntcn,
  title     = {{ModernTCN}: A Modern Pure Convolution Structure for General Time Series Analysis},
  author    = {Luo, Donghao and Wang, Xue},
  booktitle = {International Conference on Learning Representations},
  year      = {2024},
  url       = {https://proceedings.iclr.cc/paper_files/paper/2024/file/86b1437c1e4c3b3c4debff98234a67e7-Paper-Conference.pdf}
}

@inproceedings{ramaswamy2000efficient,
  title     = {Efficient Algorithms for Mining Outliers from Large Data Sets},
  author    = {Ramaswamy, Sridhar and Rastogi, Rajeev and Shim, Kyuseok},
  booktitle = {Proceedings of the 2000 ACM SIGMOD International Conference on Management of Data},
  pages     = {427--438},
  year      = {2000},
  doi       = {10.1145/342009.335437}
}

@inproceedings{scholkopf1999support,
  title     = {Support Vector Method for Novelty Detection},
  author    = {Sch{\"o}lkopf, Bernhard and Williamson, Robert C. and Smola, Alex and Shawe-Taylor, John and Platt, John},
  booktitle = {Advances in Neural Information Processing Systems},
  volume    = {12},
  year      = {1999},
  url       = {https://proceedings.neurips.cc/paper_files/paper/1999/file/8725fb777f25776ffa9076e44fcfd776-Paper.pdf}
}

@inproceedings{wu2023timesnet,
  title     = {{TimesNet}: Temporal 2D-Variation Modeling for General Time Series Analysis},
  author    = {Wu, Haixu and Hu, Tengge and Liu, Yong and Zhou, Hang and Wang, Jianmin and Long, Mingsheng},
  booktitle = {International Conference on Learning Representations},
  year      = {2023},
  url       = {https://arxiv.org/abs/2210.02186}
}

@inproceedings{wu2025catch,
  title     = {{CATCH}: Channel-Aware Multivariate Time Series Anomaly Detection via Frequency Patching},
  author    = {Wu, Xingjian and Qiu, Xiangfei and Li, Zhengyu and Wang, Yihang and Hu, Jilin and Guo, Chenjuan and Xiong, Hui and Yang, Bin},
  booktitle = {International Conference on Learning Representations},
  year      = {2025},
  url       = {https://proceedings.iclr.cc/paper_files/paper/2025/file/2b25c39788e5cf11d3541de433ebf4c0-Paper-Conference.pdf}
}

@inproceedings{qiu2025duet,
  title     = {{DUET}: Dual Clustering Enhanced Multivariate Time Series Forecasting},
  author    = {Qiu, Xiangfei and Wu, Xingjian and Lin, Yan and Guo, Chenjuan and Hu, Jilin and Yang, Bin},
  booktitle = {Proceedings of the 31st ACM SIGKDD Conference on Knowledge Discovery and Data Mining},
  pages     = {1185--1196},
  year      = {2025},
  doi       = {10.1145/3690624.3709325}
}

@article{li2027quals,
  title         = {{QUALS}: Corpus Equilibrium for Universal Forecasting via Pattern Quantization and Learnability Synchronization},
  author        = {Li, Yujie and Shao, Zezhi and Yu, Chengqing and Fu, Yisong and Zhu, Weijie and Du, Yifan and Hu, Jilin and Yang, Bin and Xu, Yongjun and Wang, Fei},
  journal       = {Proceedings of the VLDB Endowment},
  volume        = {20},
  number        = {2},
  pages         = {108--121},
  year          = {2027},
  eprint        = {2609.20156},
  archivePrefix = {arXiv},
  url           = {https://arxiv.org/abs/2609.20156}
}

@inproceedings{deng2021graph,
  title={Graph Neural Network-Based Anomaly Detection in Multivariate Time Series},
  author={Deng, Ailin and Hooi, Bryan},
  booktitle={Proceedings of the AAAI Conference on Artificial Intelligence},
  volume={35},
  number={5},
  pages={4027--4035},
  year={2021},
  doi={10.1609/aaai.v35i5.16523},
  url={https://ojs.aaai.org/index.php/AAAI/article/view/16523}
}

@inproceedings{faber2025xlstmad,
  title     = {{xLSTMAD}: A Powerful {xLSTM}-based Method for Anomaly Detection},
  author    = {Faber, Kamil and Pietro{\'n}, Marcin and {\.Z}urek, Dominik and Corizzo, Roberto},
  booktitle = {IEEE International Conference on Data Mining},
  pages     = {247--256},
  year      = {2025},
  doi       = {10.1109/ICDM65498.2025.00032}
}

@article{yeh2026matrix,
  title   = {Matrix Profile for Time-Series Anomaly Detection: A Reproducible Open-Source Benchmark on {TSB-AD}},
  author  = {Yeh, Chin-Chia Michael},
  journal = {arXiv preprint arXiv:2604.02445},
  year    = {2026},
  url     = {https://arxiv.org/abs/2604.02445}
}

@inproceedings{ozer2026streamvae,
  title     = {{STREAM-VAE}: Dual-Path Routing for Slow and Fast Dynamics in Vehicle Telemetry Anomaly Detection},
  author    = {{\"O}zer, Kadir-Kaan and Ebeling, Ren{\'e} and Enzweiler, Markus},
  booktitle = {IEEE Intelligent Vehicles Symposium},
  pages     = {444--451},
  year      = {2026},
  doi       = {10.1109/IV66570.2026.11623993}
}

@inproceedings{dutta2026giving,
  title     = {Giving Sensors a Voice: Multimodal {JEPA} for Semantic Time-Series Embeddings},
  author    = {Dutta, Utsav and Pastrana, Gerardo and Khoshfetrat Pakazad, Sina and Ohlsson, Henrik},
  booktitle = {Proceedings of the 43rd International Conference on Machine Learning},
  volume    = {306},
  series    = {Proceedings of Machine Learning Research},
  year      = {2026},
  url       = {https://arxiv.org/abs/2605.31580}
}

@inproceedings{gao2024units,
  title     = {{UniTS}: A Unified Multi-Task Time Series Model},
  author    = {Gao, Shanghua and Koker, Teddy and Queen, Owen and Hartvigsen, Thomas and Tsiligkaridis, Theodoros and Zitnik, Marinka},
  booktitle = {Advances in Neural Information Processing Systems},
  volume    = {37},
  pages     = {140589--140631},
  year      = {2024},
  doi       = {10.52202/079017-4463}
}

@inproceedings{zhou2023one,
  title     = {One Fits All: Power General Time Series Analysis by Pretrained {LM}},
  author    = {Zhou, Tian and Niu, Peisong and Wang, Xue and Sun, Liang and Jin, Rong},
  booktitle = {Advances in Neural Information Processing Systems},
  volume    = {36},
  pages     = {43322--43355},
  year      = {2023},
  doi       = {10.52202/075280-1877}
}

@inproceedings{liu2024unitime,
  title     = {{UniTime}: A Language-Empowered Unified Model for Cross-Domain Time Series Forecasting},
  author    = {Liu, Xu and Hu, Junfeng and Li, Yuan and Diao, Shizhe and Liang, Yuxuan and Hooi, Bryan and Zimmermann, Roger},
  booktitle = {Proceedings of the ACM Web Conference 2024},
  year      = {2024},
  url       = {https://arxiv.org/abs/2310.09751}
}

@inproceedings{nie2023time,
  title     = {A Time Series Is Worth 64 Words: Long-term Forecasting with Transformers},
  author    = {Nie, Yuqi and Nguyen, Nam H. and Sinthong, Phanwadee and Kalagnanam, Jayant},
  booktitle = {International Conference on Learning Representations},
  year      = {2023},
  url       = {https://arxiv.org/abs/2211.14730}
}

@article{kowsher2024llmmixer,
  title   = {{LLM-Mixer}: Multiscale Mixing in {LLM}s for Time Series Forecasting},
  author  = {Kowsher, Md and Sobuj, Md. Shohanur Islam and Prottasha, Nusrat Jahan and Alanis, E. Alejandro and Garibay, Ozlem Ozmen and Yousefi, Niloofar},
  journal = {arXiv preprint arXiv:2410.11674},
  year    = {2024},
  url     = {https://arxiv.org/abs/2410.11674}
}

@inproceedings{liu2024timer,
  title     = {{Timer}: Generative Pre-trained Transformers Are Large Time Series Models},
  author    = {Liu, Yong and Zhang, Haoran and Li, Chenyu and Huang, Xiangdong and Wang, Jianmin and Long, Mingsheng},
  booktitle = {Proceedings of the 41st International Conference on Machine Learning},
  volume    = {235},
  series    = {Proceedings of Machine Learning Research},
  pages     = {32369--32399},
  year      = {2024},
  url       = {https://proceedings.mlr.press/v235/liu24cb.html}
}

@inproceedings{ekambaram2024tiny,
  title     = {Tiny Time Mixers ({TTM}s): Fast Pre-trained Models for Enhanced Zero/Few-Shot Forecasting of Multivariate Time Series},
  author    = {Ekambaram, Vijay and Jati, Arindam and Dayama, Pankaj and Mukherjee, Sumanta and Nguyen, Nam H. and Gifford, Wesley M. and Reddy, Chandra and Kalagnanam, Jayant},
  booktitle = {Advances in Neural Information Processing Systems},
  volume    = {37},
  pages     = {74147--74181},
  year      = {2024},
  doi       = {10.52202/079017-2359}
}

@article{rasul2024lagllama,
  title   = {{Lag-Llama}: Towards Foundation Models for Probabilistic Time Series Forecasting},
  author  = {Rasul, Kashif and Ashok, Arjun and Williams, Andrew Robert and Ghonia, Hena and Bhagwatkar, Rishika and Khorasani, Arian and Darvishi Bayazi, Mohammad Javad and Adamopoulos, George and Riachi, Roland and Hassen, Nadhir and Bilo{\v{s}}, Marin and Garg, Sahil and Schneider, Anderson and Chapados, Nicolas and Drouin, Alexandre and Zantedeschi, Valentina and Nevmyvaka, Yuriy and Rish, Irina},
  journal = {arXiv preprint arXiv:2310.08278},
  year    = {2024},
  url     = {https://arxiv.org/abs/2310.08278}
}

@article{wang2025aries,
  title   = {{ARIES}: Relation Assessment and Model Recommendation for Deep Time Series Forecasting},
  author  = {Wang, Fei and Li, Yujie and Shao, Zezhi and Yu, Chengqing and Fu, Yisong and An, Zhulin and Xu, Yongjun and Cheng, Xueqi},
  journal = {arXiv preprint arXiv:2509.06060},
  year    = {2025},
  url     = {https://arxiv.org/abs/2509.06060}
}

@inproceedings{shao2025blast,
  title     = {{BLAST}: Balanced Sampling Time Series Corpus for Universal Forecasting Models},
  author    = {Shao, Zezhi and Li, Yujie and Wang, Fei and Yu, Chengqing and Fu, Yisong and Qian, Tangwen and Xu, Bin and Diao, Boyu and Xu, Yongjun and Cheng, Xueqi},
  booktitle = {Proceedings of the 31st ACM SIGKDD Conference on Knowledge Discovery and Data Mining V. 2},
  pages     = {2502--2513},
  year      = {2025},
  doi       = {10.1145/3711896.3736860}
}

@inproceedings{li2026apt,
  title     = {{APT}: Affine Prototype-Timestamp for Time Series Forecasting Under Distribution Shift},
  author    = {Li, Yujie and Shao, Zezhi and Yu, Chengqing and Fu, Yisong and Sun, Tao and Xu, Yongjun and Wang, Fei},
  booktitle = {Proceedings of the AAAI Conference on Artificial Intelligence},
  volume    = {40},
  number    = {18},
  pages     = {15180--15188},
  year      = {2026},
  doi       = {10.1609/aaai.v40i18.38542}
}

@article{aksu2024gifteval,
  title   = {{GIFT-Eval}: A Benchmark for General Time Series Forecasting Model Evaluation},
  author  = {Aksu, Taha and Woo, Gerald and Liu, Juncheng and Liu, Xu and Liu, Chenghao and Savarese, Silvio and Xiong, Caiming and Sahoo, Doyen},
  journal = {arXiv preprint arXiv:2410.10393},
  year    = {2024},
  url     = {https://arxiv.org/abs/2410.10393}
}
\bibliographystyle{iclr2027_conference}

\clearpage
\appendix
% Appendix tables use one size step above the compact main-text tables.
\newcommand{\AppendixTableStyle}{\fontsize{8}{9.2}\selectfont\renewcommand{\arraystretch}{0.68}}

\section{Notation}
\label{app:notation}

The notation follows Figure~\ref{fig:overview}: hats mark predictions,
$\mathrm{GT}$ marks realized targets, and $y_t$ denotes anomaly labels.
Uppercase $X$ and $Z$ denote sequences or hierarchical collections, and
lowercase symbols denote their elements, with token vectors in bold. Hierarchy
levels are indexed from fine to coarse, consistent with the ordering
$z_1,\ldots,z_H$ in the figure, while a superscript $(d)$ identifies the processing
direction when the same construction is applied forward and backward. Symbols
used only in a local derivation are defined where they first appear.

\begin{table}[!ht]
\centering
\caption{Notation used in Figure~\ref{fig:overview}, the method, and the appendix.}
\label{tab:notation}
\AppendixTableStyle
\renewcommand{\arraystretch}{0.765}
\begin{tabular}{>{\raggedright\arraybackslash}m{0.24\textwidth}>{\raggedright\arraybackslash}m{0.68\textwidth}}
\toprule
Symbol & Definition \\
\midrule
$t,\tau,c,h$ & Time index, target-token position, variable, and hierarchy level. \\
$j,a,d,f,k$ & Base discrepancy channel, temporal alignment, processing direction, dataset, and appended-tree index. \\
\specialrule{0.2pt}{0pt}{0pt}
$X=(x_{t,c})\in\mathbb R^{T\times C}$ & Input time series with $T$ time steps and $C$ variables; $X_{\mathrm{hist}}$ and $X_{\mathrm{GT}}$ denote the observed history and realized target segment. \\
\specialrule{0.2pt}{0pt}{0pt}
$y_t,s_t$ & Target-task anomaly label available on training sequences and the resulting task-specific anomaly score. \\
\specialrule{0.2pt}{0pt}{0pt}
$X_{\mathrm{ref}}$ & Each sequence's designated training prefix, used to estimate discrepancy location and scale. \\
\specialrule{0.2pt}{0pt}{0pt}
$H,\delta_h$ & Number of hierarchy levels and temporal support of level $h$; the evaluated configuration uses $H=6$ and $(\delta_1,\ldots,\delta_H)=(8,16,32,64,128,256)$. \\
\specialrule{0.2pt}{0pt}{0pt}
$Z_h=(\mathbf z_{\tau,h})_{\tau=1}^{N_h}$ & Continuous token sequence containing $N_h$ tokens at hierarchy level $h$; $Z=\{Z_h\}_{h=1}^{H}$ is the complete token pyramid. \\
\specialrule{0.2pt}{0pt}{0pt}
$E_\theta,\mathcal P_{\omega,h},\mathcal Q$ & Tokenizer encoding map including continuous-token projection, level-$h$ next-token predictor, and frozen codebook lookup. \\
$D_h,D_{\mathrm{full}}$ & Level-$h$ native-resolution decoder and full-resolution decoder. \\
\specialrule{0.2pt}{0pt}{0pt}
$\widehat Z_\tau,Z_{\mathrm{GT},\tau}$ & Predicted and realized target token pyramids at target position $\tau$. \\
\specialrule{0.2pt}{0pt}{0pt}
$\widehat\mu_{\tau,h},\mu_{\mathrm{GT},\tau,h}$ & Predicted and realized patch means at level $h$. \\
\specialrule{0.2pt}{0pt}{0pt}
$\widehat\sigma^2_{\tau,h},\sigma^2_{\mathrm{GT},\tau,h}$ & Predicted and realized patch variances. Their logarithms are $\widehat\ell_{\tau,h}$ and $\ell_{\mathrm{GT},\tau,h}$. \\
\specialrule{0.2pt}{0pt}{0pt}
$\widehat X,X_{\mathrm{GT}}$ & Decoded prediction and realized target in observation space. \\
\specialrule{0.2pt}{0pt}{0pt}
$R_h,R_{t,h}^{(d)}$ & Level-$h$ risk evidence in Figure~\ref{fig:overview} and its time- and direction-indexed form. \\
$\mathbf R_t^{(d)}$ & Complete structured risk representation for $d\in\{\rightarrow,\leftarrow\}$. \\
\specialrule{0.2pt}{0pt}{0pt}
$\mathcal E_{t,c}^{(d)},e_{t,c,j}^{(d)}$ & Set of 64 base discrepancy channels indexed by emission time $t$ and its channel-$j$ component for variable $c$. \\
$\widetilde e_{t,c,j}^{(d)}$ & Held, log-transformed value of the base discrepancy. \\
\specialrule{0.2pt}{0pt}{0pt}
$\mathcal C(\cdot;m,\rho)$ & Robust calibration with reference median $m$ and interquartile range $\rho$. \\
\specialrule{0.2pt}{0pt}{0pt}
$F_B,F_f$ & Track-specific base ensemble and selected dataset ensemble. \\
$K_f,\Delta F_{f,k}$ & Number of appended trees and the $k$th appended tree for dataset $f$. \\
\specialrule{0.2pt}{0pt}{0pt}
$\widetilde{s}_t^{(d)},s_t$ & Direction-specific and final bidirectional anomaly scores. \\
\bottomrule
\end{tabular}
\end{table}

\FloatBarrier

\section{Method Details}
\label{app:method-details}

\subsection{Exact Predictive-Evidence Definitions}
\label{app:exact-evidence}

For a fixed direction $d$, target position $\tau$, and level $h$, abbreviate the
predicted and realized target tokens $\widehat{\mathbf z}_{\tau,h}$ and
$\mathbf z_{\mathrm{GT},\tau,h}$ as $\widehat{\mathbf z}$ and
$\mathbf z_{\mathrm{GT}}$. We suppress the direction superscript throughout
this subsection. With
$\widehat{\mathbf u}=\widehat{\mathbf z}/\|\widehat{\mathbf z}\|$ and
$\mathbf u_{\mathrm{GT}}=\mathbf z_{\mathrm{GT}}/\|\mathbf z_{\mathrm{GT}}\|$,
HieraNTP removes the
radial component of their difference and evaluates the tangent residual under
the level-specific precision matrix $\mathbf P_h$:
\begin{equation}
\mathbf r_Z=(\mathbf I-\mathbf u_{\mathrm{GT}}\mathbf u_{\mathrm{GT}}^\top)
(\widehat{\mathbf u}-\mathbf u_{\mathrm{GT}}),
\qquad
e_Z=\mathbf r_Z^\top\mathbf P_h\mathbf r_Z.
\label{eq:retrieval-evidence}
\end{equation}
This score measures the tangent component relative to the realized token direction.
With identity precision, it equals one minus the squared cosine similarity
between the two tokens.
Patch-mean and patch-variance discrepancies use
$e_\mu=\operatorname{SmoothL1}(\widehat\mu,\mu_{\mathrm{GT}})$ and
$e_\ell=\operatorname{SmoothL1}(\widehat\ell,\ell_{\mathrm{GT}})$, where
$\widehat\ell=\log\widehat\sigma^2$ and
$\ell_{\mathrm{GT}}=\log\sigma_{\mathrm{GT}}^2$. The two temporal alignments
offset the token grid by 0 and 4 observations before predictive inference.
Each variable has $2\times6\times3=\HieraPredDim$ token and
local-statistic channels across the six levels.

Before decoding, the frozen codebook lookup $\mathcal Q$ maps a continuous
token to its corresponding quantized decoder state. The level-specific decoder
$D_h$ then supplies two additional comparisons,
\begin{equation}
e_{h,X}=|D_h(\mathcal Q(\widehat{\mathbf z}))-X_{\mathrm{GT},h}|,
\qquad
e_{h,Z}=|D_h(\mathcal Q(\widehat{\mathbf z}))
-D_h(\mathcal Q(\mathbf z_{\mathrm{GT}}))|,
\label{eq:scale-decoded-evidence}
\end{equation}
where $X_{\mathrm{GT},h}$ is the realized observation target at level $h$.
The two offsets and six levels yield
$2\times6\times2=\HieraScaleDecodedDim$ channels. The
full-resolution decoder operates on the complete predicted and realized token
pyramids $\widehat Z_\tau$ and $Z_{\mathrm{GT},\tau}$, with $\mathcal Q$
applied to each token, and provides
\begin{equation}
e_{\mathrm{full},X}(t)=|D_{\mathrm{full}}(\mathcal Q(\widehat Z_\tau))_t-X_{\mathrm{GT},t}|,
\qquad
e_{\mathrm{full},Z}(t)=
|D_{\mathrm{full}}(\mathcal Q(\widehat Z_\tau))_t
-D_{\mathrm{full}}(\mathcal Q(Z_{\mathrm{GT},\tau}))_t|.
\label{eq:full-decoded-evidence}
\end{equation}
The full-resolution comparisons complete the
$\HieraPredDim+\HieraScaleDecodedDim+\HieraFullDecodedDim=\HieraBaseDim$
channels per variable.

\subsection{Direct Evidence Construction}
\label{app:direct-evidence}

Each channel uses robust center and scale statistics from finite
forward-direction emissions in the designated reference prefix. The center is
the median, and the initial scale is $1.4826$ times the median absolute
deviation. Degenerate scales fall back successively to the interquartile range
divided by $1.349$, the standard deviation, and one. Standardized emissions are
projected over their target spans and averaged where spans overlap. The four
full-resolution channels remain pointwise. The patch-mean discrepancy uses excess
threshold 2, while the remaining channels use threshold 1. Positive excess is
clipped to $[0,50]$ and transformed by $\log(1+x)$.

Variable means and maxima over valid variables yield 128 multivariate direct
channels, while univariate sequences retain 64. Every pooled channel
then undergoes an average-tie rank transform over the complete sequence.
The precision matrices in Eq.~\ref{eq:retrieval-evidence} are estimated
separately for TSB-AD-U and TSB-AD-M from forward, offset-0 residuals whose
target spans lie within the designated Tuning prefixes. Each sequence supplies
at most 512 evenly spaced residuals per level. Ledoit--Wolf shrinkage
\citep{ledoit2004wellconditioned} estimates
the covariance around the sample mean, and its inverse weights the uncentered
tangent residual during scoring. Prefix boundaries follow file metadata
(\texttt{tr}), and fitting uses all eligible residuals irrespective of labels.
Each track's matrices are shared across alignments and directions and reused
for the corresponding TAB setting.

\subsection{Reference Calibration}
\label{app:reference-evidence}

Sparse discrepancy emissions are held until the next valid update and transformed as
\begin{equation}
\widetilde e_{t,c,j}^{(d)}=\log\!\left(
1+\max(e_{\pi_{d,c,j}(t),c,j}^{(d)},0)\right),
\label{eq:held-error}
\end{equation}
where $j\in\{1,\ldots,\HieraBaseDim\}$ and the locally defined map
$\pi_{d,c,j}(t)$ returns the latest valid emission time for variable $c$ and
channel $j$ at or before $t$ in direction $d$.
Given reference median $m$ and interquartile range $\rho$, the robust calibration is
\begin{equation}
\mathcal C(x;m,\rho)=\operatorname{asinh}\!\left(
\frac{x-m}{\max(\rho,0.05)}\right).
\label{eq:normal-calibration}
\end{equation}
The pool-then-calibrate branch first aggregates variables by their mean and
maximum and estimates $(m,\rho)$ from the final 40\% of the reference prefix. The
calibrate-then-pool branch estimates the same statistics for each variable
before applying variable mean and maximum pooling.

\paragraph{Support rules.}

Variable-relative calibration requires both the current held value and the value
32 steps earlier to be valid. Each channel becomes active only if the first 60\% of the
reference prefix contains at least two fresh emissions and 64 support-eligible held
values, and the remaining 40\% contains at least one valid value. Its median and
interquartile range are then estimated from that remaining 40\%.
Forward-direction reference statistics are applied unchanged to both directions.

\subsection{Temporal Summaries}
\label{app:temporal-evidence}

For each instantaneous channel $u_t$, the temporal summaries are
\begin{equation}
\mathcal T(u)_t=\left[u_t;
\operatorname{mean}_{L}(u)_t;
\operatorname{max}_{L}(u)_t:
L\in\mathcal L_{\mathrm{win}}\right],
\qquad
\mathcal L_{\mathrm{win}}=\{9,17,33,65,129,257\}.
\label{eq:temporal-bank}
\end{equation}
Centered windows use nearest-value extension at sequence boundaries. The current
value, six means, and six maxima give $\HieraTemporalOps$ values per evidence
channel.

\begin{table}[t]
\centering
\caption{Composition of the 64 base discrepancy channels.}
\label{tab:base-channel-composition}
\AppendixTableStyle
\renewcommand{\arraystretch}{0.765}
\begin{tabularx}{\textwidth}{@{}lXc@{}}
\toprule
Source & Comparison at each token-grid offset & Channels \\
\midrule
Token-space & Predicted and realized tokens at six levels & $2\times6=12$ \\
Patch mean & Predicted and realized patch means at six levels & $2\times6=12$ \\
Patch variance & Predicted and realized patch variances at six levels & $2\times6=12$ \\
Decoded observation & Decoded prediction and observed target at six native scales and full resolution & $2\times(6+1)=14$ \\
Decoded target & Decoded prediction and realized-token reconstruction at six native scales and full resolution & $2\times(6+1)=14$ \\
\midrule
Total & Two token-grid offsets: 0 and 4 observations & 64 \\
\bottomrule
\end{tabularx}
\end{table}

\begin{table}[!ht]
\centering
\caption{Processing paths and resulting representation dimensions.}
\label{tab:appendix-dimensions}
\AppendixTableStyle
\renewcommand{\arraystretch}{0.765}
\begin{tabularx}{\textwidth}{@{}lXrr@{}}
\toprule
Variant & Ordered operations before temporal summaries & U & M \\
\midrule
Direct & Robust standardization, target-interval projection, positive-excess log transform, variable aggregation, sequence ranking & 832 & 1664 \\
Pool-then-calibrate & Hold and log-transform responses, aggregate variables, calibrate the aggregate & 832 & 1664 \\
Calibrate-then-pool & Hold and log-transform responses, calibrate eligible variable responses, aggregate valid variables & 832 & 1664 \\
\midrule
Complete representation & Concatenation of the three paths after temporal summaries & 2496 & 4992 \\
\bottomrule
\end{tabularx}
\end{table}

\subsection{Risk-Evidence Construction}
\label{app:representation-dimensions}

The comparisons in Appendix~\ref{app:exact-evidence} yield 64 base channels
per variable and processing direction, as enumerated in
Table~\ref{tab:base-channel-composition}. Each channel retains its source,
temporal support, and token-grid offset throughout subsequent processing.
The reference prefix is the same sequence's training segment, delimited by
\texttt{tr} in TSB-AD and \texttt{train\_lens} in TAB.

\paragraph{Time alignment.}
For Direct evidence, standardized patch responses are assigned to every
observation in the predicted patch and averaged where target intervals overlap.
The four full-resolution channels retain their pointwise responses.
Patch emissions occur at the target's right boundary in forward processing
and left boundary in reverse processing. The calibrated variants carry each
valid emission until the next update, with held responses initialized to zero.
Both constructions preserve separate
channels for each support and offset on the original observation grid.

\paragraph{Processing order.}
Table~\ref{tab:appendix-dimensions} summarizes the three paths using the
transformations defined in Appendices~\ref{app:direct-evidence}--\ref{app:temporal-evidence}.
Reference statistics are estimated from forward-direction training responses
and reused in both directions. In univariate sequences, variable aggregation
is the identity. Calibrate-then-pool retains its support filtering, whereas
Pool-then-calibrate uses all held reference responses, so the paths can
still differ.

Applying the 13 temporal summaries to each path yields
$64\times13=832$ univariate features.
Multivariate sequences retain both variable means and maxima, doubling this
count. Concatenation yields 2496 and 4992 features, respectively, for each
direction's input to the detector.

\subsection{Task-Adaptive Detector Definition}
\label{app:detector-definition}

Let $F_B$ denote the additive function of the track-specific base ensemble. For
a dataset $f$, the selected ensemble is
\begin{equation}
F_f(\mathbf R)=F_B(\mathbf R)+
\sum_{k=1}^{K_f}\Delta F_{f,k}(\mathbf R),
\label{eq:base-detector}
\end{equation}
where $K_f=0$ and $F_f=F_B$ when no extension is used. TAB uses $K_f=32$
for adapted datasets, while TSB-AD uses the dataset-specific settings in
Appendix~\ref{app:detector-parameters}. For each processing
direction $d$, this ensemble produces
\begin{equation}
\widetilde{s}_t^{(d)}=\operatorname{sigmoid}\!\left(F_f(\mathbf R_t^{(d)})\right).
\label{eq:family-adaptation}
\end{equation}
The base trees remain fixed while the new trees are appended, so each dataset
model is a single complete ensemble. After aligning the two processing
directions, the final score is
\begin{equation}
s_t=\tfrac12\left(\widetilde{s}_t^{(\rightarrow)}+\widetilde{s}_t^{(\leftarrow)}\right).
\label{eq:bidirectional-score}
\end{equation}

\FloatBarrier

\section{Training Data and Model Configuration}
\label{app:implementation}

\subsection{Tokenizer and Next-Token Predictor Pretraining}
\label{app:upstream-pretraining}

\paragraph{Multiscale tokenizer pretraining.}
The causal tokenizer uses a patch size and stride of 8, six Swin stages with
depths $(2,2,2,2,2,2)$, attention-head counts $(6,6,8,12,16,16)$, and
384-dimensional encoder states. It uses rotary position embeddings, and its
local attention window contains the current position and the preceding 15 positions. In the
paper notation, levels $h=1,\ldots,H$ are ordered from fine to coarse and have
temporal supports
$(\delta_1,\ldots,\delta_H)=(8,16,32,64,128,256)$. Every encoder state is projected to a 48-dimensional
continuous token, and all levels share an $8192\times48$
vector-quantization codebook. The decoder uses the coarsest quantized sequence
as its query and the complete token pyramid as multiscale key-value memory.
Full-resolution and native-scale reconstruction losses use a coverage-weighted
average of absolute and squared error. The multiscale reconstruction weight
is 0.3, and the vector-quantization commitment weight is 0.5. The tokenizer
is trained for 200,000 steps on the QUALS-organized GIFT-Eval Pretrain
corpus \citep{li2027quals}, followed by
20,000 additional steps on QUALS. The two stages use initial learning rates
$10^{-4}$ and $10^{-5}$, respectively, with cosine decay and warmup.
Training uses AdamW with $\beta=(0.9,0.95)$, weight decay 0.01, gradient
clipping at 1, and an effective batch size of 336 across seven GPUs.
The evaluated checkpoint is \texttt{Swin\_20000.pt}, with 47,512,382
encoder--decoder parameters.

\paragraph{Hierarchical next-token pretraining.}
The hierarchical next-token predictor $\mathcal P_{\omega,h}$ has hidden width
256, four attention heads, and feed-forward width 1024. The finest level contains four causal
self-attention blocks. Each coarser level aggregates temporally overlapping
child states and applies same-scale self-attention followed by cross-attention
to the corresponding frozen tokenizer memory. Attention permits a key only when
its raw endpoint does not exceed the target start. Dropout is 0.1. Rotary
position embeddings use base 10,000 and position scale 8. The feedback bridge
adds a $48\to128\to384$ residual multilayer perceptron to a $48\to384$
linear map to convert predicted tokens into encoder states.
Input projection and feedback parameters are shared across
levels, whereas attention blocks and prediction heads are level-specific.

The staged predictor curriculum starts with entirely synthetic regular
series and increases the QUALS training-data proportion in 20-percentage-point
increments at 20,000-step intervals. The evaluated checkpoint,
\texttt{msi\_blt\_step\_013000.pt}, is saved after 13,000 steps with an
80\% QUALS and 20\% synthetic mixture, initialized from the preceding
stage's step-80,000 checkpoint. The synthetic series used in this stage
contain no injected anomalous events or additive noise.
Training samples
contain 512--1024 observed points followed by at least 1024 target points, with a
maximum total length of 4096. Normalization statistics are computed only from
the observed history. Teacher forcing and detached rollout are used in equal
proportions. The final stage uses AdamW with constant learning rates
$3\times10^{-5}$ for the predictor and $9\times10^{-5}$ for the feedback
bridge, weight decay 0.01, gradient clipping at 1, and mixed precision.
The effective batch size is 320 across four GPUs, with no learning-rate warmup.
The loss is
\begin{equation}
  \mathcal L_{\mathrm{NTP}}
  =\mathcal L_{\mathrm{token}}
  +0.1\mathcal L_{\mathrm{mean}}
  +0.1\mathcal L_{\mathrm{logvar}}
  +0.1\mathcal L_{\mathrm{feedback}},
  \label{eq:ntp-training-objective}
\end{equation}
where token and feedback consistency use cosine loss and the two local-statistic
terms use Smooth-L1 loss. Level-balanced weighting uses exponent 0.25 and caps
individual level weights at 2.0. The predictor learns continuous token targets
with codebook-classification loss disabled. All pretrained components remain
frozen during anomaly detection.

\paragraph{Synthetic sequences.}
The final-stage generator combines periodic patterns and trends:
\begin{equation}
  x_t^{\mathrm{syn}}
  =\frac{w_P}{s(P^*)}P_t^*+\frac{w_T}{s(B)}B_t,
  \qquad s(v)=\max\{\operatorname{std}(v),10^{-8}\}.
  \label{eq:synthetic-composition}
\end{equation}
$P_t^*$ allows gain, envelope, phase, and within-cycle shape variations
of a periodic pattern, and $B_t$ is a trend. Table~\ref{tab:synthetic-patterns}
summarizes the forms and sampling strata. Outside the strong-trend stratum,
$w_P\sim\mathcal U(0.75,1)$ and a trend is included with probability 0.45;
when included, $w_T/w_P\sim\mathcal U(0.08,0.55)$, and otherwise $w_T=0$.
Periods are sampled approximately log-uniformly within each interval, with
an off-grid preference of 0.8 relative to the 8-point token grid.
The table reports proportions within the synthetic stream; eight of ten
equal-sized batches use QUALS, and two are synthetic.

\begin{table}[!ht]
\centering
\caption{Synthetic patterns and sampling strata for next-token pretraining.
Component formulas are representative forms before amplitude scaling and
normal shape variations; sampling proportions apply to the four strata.}
\label{tab:synthetic-patterns}
\AppendixTableStyle
\setlength{\tabcolsep}{5pt}
\begin{tabularx}{\textwidth}{@{}lXc@{}}
\toprule
Name & Generating form or sampling rule & Share \\
\midrule
Sinusoidal & $P_t=\sin\theta_t$ & --- \\
Multiharmonic & $P_t=\sum_{k=1}^{K}a_k\sin(k\theta_t+\phi_k)$ & --- \\
Square & $P_t=\operatorname{sign}(\sin\theta_t)$ & --- \\
Triangular & $P_t=2|2r_t|-1$ & --- \\
Sawtooth & $P_t=2r_t$ & --- \\
Repeated motif & $P_t=m(\operatorname{frac}(v_t))$; smooth piecewise patterns or pulse trains & --- \\
Smooth trend & $B_t\in\{at,\,e^{100\beta u_t},\,\log(1+t),\,(u_t+1/L)^\alpha\}$ & --- \\
\midrule
General & $p\in[7,1536]$; optional trend & 20\% \\
Strong trend & $p\in[7,1536]$; $(w_P,w_T)=(0.55,1)$, trend contrast $\geq0.75$ & 30\% \\
Long period & $p\in[512,1023]$ or $[1024,1536]$, equally weighted; optional trend & 25\% \\
Short period & $p\in[7,31]$; optional trend & 25\% \\
\bottomrule
\end{tabularx}
\begin{minipage}{\textwidth}
\AppendixTableStyle
For a length-$L$ parent sequence, $u_t=t/(L-1)$,
$v_t=(t+\phi)/p$, $\theta_t=2\pi v_t$, and
$r_t=v_t-\lfloor v_t+1/2\rfloor$; $m$ is a repeating within-cycle shape.
The general and strong-trend strata allocate 15/20/25/20/20\% to period
intervals 7--31/32--127/128--511/512--1023/1024--1536.
Trend contrast is the absolute difference between the last- and first-quarter
means divided by the sequence standard deviation.
\end{minipage}
\end{table}

\paragraph{Predictive inference.}
Detection uses 4096-observation contexts and 1024-observation chunks with
672-observation overlap, giving a stride of 352. Input normalization fixes the mean and
standard deviation from the designated sequence prefix. Extraction runs in
both temporal directions with token-grid offsets of 0 and 4 observations.
Repeated records are identified by view, level, and global target span.
Forward and reverse emissions are timestamped at the last and first target
observations in the original sequence, respectively.
Temporal summaries extend endpoint values.

\subsection{Task-Adaptive Detector Parameters}
\label{app:detector-parameters}

\begin{table}[!ht]
\centering
\caption{Initial-stage XGBoost parameters. TAB uses the same parameters for
both tracks and appends 32 trees for each adapted dataset.}
\label{tab:xgb-parameters}
\AppendixTableStyle
\begin{tabular}{lcccc}
\toprule
Parameter & TSB-AD-U & TSB-AD-M & TAB univariate & TAB multivariate \\
\midrule
Feature dimension & 2496 & 4992 & 2496 & 4992 \\
Number of trees & 384 & 384 & 384 & 384 \\
Maximum depth & 3 & 6 & 6 & 6 \\
Minimum child weight & 64 & 4 & 4 & 4 \\
$L_2$ regularization & 10 & 100 & 100 & 100 \\
Column subsampling & 0.35 & 0.75 & 0.75 & 0.75 \\
Learning rate & 0.025 & 0.08 & 0.08 & 0.08 \\
Row subsampling & 0.7 & 0.7 & 0.7 & 0.7 \\
\bottomrule
\end{tabular}
\end{table}

All four configurations use GPU XGBoost 2.1.4 with binary logistic loss,
256-bin histogram tree construction, seed 20260930, and zero initial-stage
$L_1$ regularization.

For TSB-AD-M, dataset-specific adaptation selects among no extension and
32 or 96 trees of depth 2 or 4. Appended trees use learning rate 0.03,
minimum child weight 16, $L_2$ regularization 30, zero $L_1$ regularization,
row subsampling 0.85, and column subsampling 0.75.
TSB-AD-U uses an expanded search with 24 parameter configurations and
32, 96, 192, or 384 appended trees. The sampled parameter ranges are depth
$\{2,3,4,6\}$, learning rate $\{0.015,0.03,0.06,0.1\}$,
minimum child weight $\{4,16,32,64\}$, $L_2$ regularization
$\{3,10,30,100\}$, $L_1$ regularization $\{0,0.1,1\}$, row
subsampling $\{0.7,0.85,1\}$, and column subsampling $\{0.5,0.75,1\}$.
Two configurations use the fixed depth-2 and depth-4 recipes above; the
remaining 22 are sampled with seed 20260930. The resulting detectors append
trees for 17 U and five M datasets. At evaluation, the dataset identifies
the corresponding complete ensemble; other datasets use the shared base.

TSB-AD base-parameter selection and appended-tree adaptation use the
held-sequence macro average of $(\mathrm{VUS\mbox{-}PR}+\mathrm{AP})/2$.
The M protocol requires at least two eligible held-out sequences and both
classes in each dataset-training fold. The U search uses two
dataset-stratified sequence folds when multiple Tuning sequences are available.
Singleton datasets use a temporal holdout with a 4224-point exclusion on
each side of the boundary when both partitions contain at least eight points
of each class and 512 timestamps. Otherwise, they use the configuration
selected by pooled validation on other Tuning datasets. The detector is fitted
on sampled timestamps whose indices, labels, and fixed weights are recorded
for every Tuning sequence in the protocol artifacts.

\paragraph{TAB detector training.}
\label{app:tab-detector-training}
% Verified against FAMILY32_PARAMETERS.json, FAMILY32_TRAINING_CONFIG_CN.md,
% and 55_run_ntp64_base384_family_append32.py in
% OpenTS_BASE384_FAMILY32_paper_bundle_20260925.zip.
The univariate and multivariate tracks each train a shared base detector from scratch,
using the training prefixes of 1635 and 220 sequences across 15 and 29
datasets, respectively. Both tracks use the initial-stage parameters in
Table~\ref{tab:xgb-parameters}, producing 384 base trees. Each dataset
whose sampled training rows contain both normal and anomalous observations
receives an independent 32-tree extension initialized from the same base
ensemble. The original 384 trees remain unchanged. These appended trees have maximum depth 2,
learning rate 0.03, minimum child weight 16, $L_2$ regularization 30,
zero $L_1$ regularization, row subsampling 0.85, and column subsampling 0.75.
The exported models contain effective dataset extensions for eight univariate
and ten multivariate datasets, yielding 416 trees per adapted ensemble.
Datasets with single-class training rows or no splits in the appended trees
use the shared 384-tree base. This applies to seven univariate datasets and
19 multivariate datasets; five of the latter meet the two-class criterion
but produce no appended-tree splits. Routing follows the dataset identity
and this training-stage rule. Both stages use fixed boosting rounds and
hyperparameters shared across datasets.

Offline evidence construction uses the complete observation sequence, including
the training prefix and test interval, with labels supplied separately.
Training rows and their labels are then selected exclusively from the training
prefix.

Normal points are sampled uniformly without replacement from each training
prefix, with caps of 64 and 512 per sequence for the univariate and multivariate
tracks. Anomalous points are sampled at evenly spaced positions within each
labeled training event, with respective caps of 32 and 256 per event.
Forward and reverse representations contribute equally weighted training rows.
The resulting matrices contain 301,888 rows with 92,608 positive labels for
the univariate track, and 302,286 rows with 77,006 positive labels for the
multivariate track.

Sample weights assign equal total mass to each dataset. For datasets containing
training anomalies, half of this mass is assigned to normal samples and half
to anomalous events. The latter is divided equally across events and then
across sampled points within each event. Datasets with normal-only training
prefixes assign their full mass to normal samples. Each temporal direction
receives half of its point's weight. Before fitting, weights are rescaled to
have mean one. For a dataset extension, rescaling applies to that dataset's
rows and preserves relative class, event, and direction weights.
The global positive-class multiplier remains one. The runs use
Python 3.9.24 and CUDA-enabled XGBoost 2.1.4 on NVIDIA RTX 4090 GPUs.

\paragraph{Implementation artifacts.}
The released downstream package begins from exported frozen HieraNTP records,
prediction evidence, and decoder outputs. It contains feature construction,
trained tree models, archived scores, and evaluation artifacts. Upstream
tokenizer and predictor training code and checkpoints are maintained separately.

\FloatBarrier

\section{Benchmarks, Baselines, and Detailed Results}
\label{app:main-experiments}

\subsection{Datasets and Supervised Evaluation}
HieraNTP is evaluated with target-task supervision: the downstream detector
uses TSB-AD Tuning labels or TAB training-prefix labels, while its tokenizer
and next-token predictor remain frozen.
TSB-AD-U provides 48 Tuning and 350 Eval sequences, and TSB-AD-M provides
20 Tuning and 180 Eval sequences \citep{liu2024tsbad}.
The Eval collections cover 23 univariate and 17 multivariate source datasets.
They include service monitoring (IOPS, NAB, and YAHOO), spacecraft telemetry
(MSL and SMAP), industrial monitoring (SWaT), and physiological signals
(LTDB, MITDB, and SVDB). Tables~\ref{tab:u-per-dataset}
and~\ref{tab:m-per-dataset} list every dataset and its Eval sequence count.

TAB uses the official full-shot training/test splits for 15 univariate
and 29 multivariate datasets \citep{qiu2025tab}.
Complete results, including OPP, appear in Tables~\ref{tab:tab-datasets-u}
and~\ref{tab:tab-datasets-m}. The benchmarks share some source datasets
but use different splits and aggregation rules.

\subsection{Comparison Methods}
The TSB-AD comparison includes leading public leaderboard submissions and
established reference methods. These cover statistical and distance-based
detectors, reconstruction and forecasting networks, and pretrained
time-series models \citep{liu2024tsbad}. Representative neural detectors
include Donut, OmniAnomaly, USAD, and TranAD
\citep{xu2018donut,su2019omnianomaly,audibert2020usad,tuli2022tranad}.
Pretrained baselines include MOMENT, Chronos, TimesFM, Lag-Llama, TimeRCD,
and TSPulse
\citep{pmlr-v235-goswami24a,ansari2024chronos,pmlr-v235-das24c,rasul2024lagllama,lan2025timercd,ekambaram2025tspulse}.
Recent submissions include xLSTMAD, MMPAD, StreamVAE, and CHARM
\citep{faber2025xlstmad,yeh2026matrix,ozer2026streamvae,dutta2026giving}.
Their detection configurations follow the TSB-AD implementations and
leaderboard releases.\footnote{\url{https://github.com/TheDatumOrg/TSB-AD}}
CHARM uses pretrained embeddings with cosine-neighbor scoring in TSB-AD.
TimeRCD+MAFT is reported under its leaderboard submission name.
Table~\ref{tab:timercd-full} reports the TimeRCD selection of 11 univariate
and five multivariate datasets using its four-metric implementation.

The TAB main table combines leading OpenTS methods with deep-learning
baselines. KMeans, KNN, EIF, and OCSVM provide clustering, distance, isolation, and
one-class references. DLinear, iTransformer, and TimesNet represent
forecasting architectures used for detection
\citep{zeng2023transformers,liu2024itransformer,wu2023timesnet}.
TranAD, DCdetector, and DualTF provide anomaly-detection comparisons
\citep{tuli2022tranad,yang2023dcdetector,nam2024breaking}.
The extended tables also include UniTS and PatchTST
\citep{gao2024units,nie2023time}, language-model adaptations GPT4TS,
UniTime, and LLMMixer \citep{zhou2023one,liu2024unitime,kowsher2024llmmixer},
and pretrained Timer and TTM \citep{liu2024timer,ekambaram2024tiny}.
The detection adaptations and scores follow TAB's OpenTS implementations
\citep{qiu2025tab}.\footnote{\url{https://github.com/decisionintelligence/TAB}}
The extended tables cover the OpenTS baselines, with missing metrics left blank.

\subsection{Evaluation Metrics}
AP summarizes precision--recall performance across score thresholds and
corresponds to the AUC-PR field computed by both benchmark implementations.
AUC-ROC summarizes the true-positive/false-positive trade-off.
VUS-PR and VUS-ROC additionally integrate over temporal tolerance
\citep{paparrizos2022vus}.
Standard-F1 evaluates pointwise predictions, range-based F1 measures
interval detection, and Affiliation-F evaluates local event correspondence
\citep{huet2022affiliation}. The TimeRCD comparison retains its F1-T
implementation and metric names \citep{lan2025timercd}.
TSB-AD results average sequence-level metrics, while the TAB leaderboard
comparison averages dataset-level metrics.
The TAB threshold rule and medal aggregation are specified in
Appendix~\ref{app:opents-protocol}.

\paragraph{TSB-AD score evaluation.}

VUS uses 250 thresholds. The tolerance window is estimated from the first
channel by autocorrelation, using at most the first 20,000 observations and 400
lags; values above 300 and failed estimates fall back to 125. Multivariate
channels are not flattened. Scores are evaluated at their archived scale
without complete-sequence min--max normalization.

\subsection{Extended TSB-AD Leaderboard Results}
\label{app:updated-leaderboard}

Tables~\ref{tab:updated-u-metrics} and~\ref{tab:updated-m-metrics} consolidate
public TSB-AD results. Recent submissions use three decimals, and the original
leaderboard retains its published two-decimal precision. Entries absent from
the public reports are left blank. For HieraNTP, the benchmark's AUC-PR field
uses \texttt{average\_precision\_score}, reported as AP.
Uncomputed F1 metrics remain blank.

\makeatletter
\setlength{\@fptop}{0pt}
\setlength{\@fpsep}{12pt}
\makeatother

\begin{table}[!htp]
\centering
\caption{Results on TSB-AD-U (350 evaluation sequences).}
\label{tab:updated-u-metrics}
\AppendixTableStyle
\setlength{\tabcolsep}{3.0pt}
\begin{tabular}{lrrrrrrrrr}
\toprule
Method & AUC-PR & AUC-ROC & VUS-PR & VUS-ROC & Std.-F1 & PA-F1 & Event-F1 & R-F1 & Aff.-F \\
\midrule
HieraNTP & \textbf{0.643} & \textbf{0.925} & \textbf{0.656} & \textbf{0.939} & -- & -- & -- & -- & -- \\
Time-RCD+MAFT (FT) & 0.590 & 0.901 & 0.586 & 0.920 & 0.555 & 0.701 & -- & -- & 0.915 \\
TSPulse (FT) & -- & -- & 0.552 & -- & -- & -- & -- & -- & -- \\
TSPulse (ZS) & -- & -- & 0.480 & -- & -- & -- & -- & -- & -- \\
CHARM & -- & -- & 0.501 & -- & -- & -- & -- & -- & -- \\
Time-RCD & 0.495 & 0.843 & 0.522 & 0.871 & 0.527 & 0.672 & -- & -- & 0.886 \\
StreamVAE & -- & -- & 0.451 & -- & -- & -- & -- & -- & -- \\
MMPAD & 0.403 & 0.783 & 0.440 & 0.787 & 0.447 & 0.592 & 0.503 & 0.435 & 0.850 \\
xLSTMAD & 0.343 & 0.761 & 0.398 & 0.829 & 0.405 & 0.604 & 0.507 & 0.353 & 0.856 \\
Sub-PCA & 0.37 & 0.71 & 0.42 & 0.76 & 0.42 & 0.56 & 0.49 & 0.41 & 0.85 \\
KShapeAD & 0.35 & 0.74 & 0.40 & 0.76 & 0.39 & 0.58 & 0.46 & 0.40 & 0.83 \\
POLY & 0.31 & 0.73 & 0.39 & 0.76 & 0.37 & 0.53 & 0.45 & 0.35 & 0.85 \\
Series2Graph & 0.33 & 0.76 & 0.39 & 0.80 & 0.38 & 0.65 & 0.50 & 0.35 & 0.85 \\
MOMENT (FT) & 0.30 & 0.69 & 0.39 & 0.76 & 0.35 & 0.65 & 0.49 & 0.35 & 0.86 \\
MOMENT (ZS) & 0.30 & 0.68 & 0.38 & 0.75 & 0.35 & 0.61 & 0.49 & 0.36 & 0.86 \\
KMeansAD & 0.32 & 0.74 & 0.37 & 0.76 & 0.37 & 0.56 & 0.44 & 0.38 & 0.82 \\
USAD & 0.32 & 0.66 & 0.36 & 0.71 & 0.37 & 0.50 & 0.43 & 0.40 & 0.84 \\
Sub-KNN & 0.27 & 0.76 & 0.35 & 0.79 & 0.34 & 0.61 & 0.43 & 0.32 & 0.84 \\
MatrixProfile & 0.26 & 0.73 & 0.35 & 0.76 & 0.33 & 0.63 & 0.44 & 0.32 & 0.84 \\
SAND & 0.29 & 0.73 & 0.34 & 0.76 & 0.35 & 0.56 & 0.42 & 0.36 & 0.81 \\
CNN & 0.33 & 0.71 & 0.34 & 0.79 & 0.38 & 0.78 & 0.66 & 0.35 & 0.88 \\
LSTMAD & 0.31 & 0.68 & 0.33 & 0.76 & 0.37 & 0.71 & 0.59 & 0.34 & 0.86 \\
SR & 0.32 & 0.74 & 0.32 & 0.81 & 0.38 & 0.87 & 0.67 & 0.35 & 0.89 \\
TimesFM & 0.28 & 0.67 & 0.30 & 0.74 & 0.34 & 0.84 & 0.63 & 0.34 & 0.89 \\
IForest & 0.29 & 0.71 & 0.30 & 0.78 & 0.35 & 0.73 & 0.56 & 0.30 & 0.84 \\
OmniAnomaly & 0.27 & 0.65 & 0.29 & 0.72 & 0.31 & 0.59 & 0.46 & 0.29 & 0.83 \\
Lag-Llama & 0.25 & 0.65 & 0.27 & 0.72 & 0.30 & 0.77 & 0.59 & 0.31 & 0.88 \\
Chronos & 0.26 & 0.66 & 0.27 & 0.73 & 0.32 & 0.83 & 0.61 & 0.33 & 0.88 \\
TimesNet & 0.18 & 0.61 & 0.26 & 0.72 & 0.24 & 0.67 & 0.47 & 0.21 & 0.86 \\
AutoEncoder & 0.19 & 0.63 & 0.26 & 0.69 & 0.25 & 0.54 & 0.36 & 0.28 & 0.82 \\
TranAD & 0.20 & 0.57 & 0.26 & 0.68 & 0.25 & 0.58 & 0.43 & 0.25 & 0.83 \\
FITS & 0.17 & 0.61 & 0.26 & 0.73 & 0.23 & 0.65 & 0.42 & 0.20 & 0.86 \\
Sub-LOF & 0.16 & 0.68 & 0.25 & 0.73 & 0.24 & 0.57 & 0.35 & 0.25 & 0.82 \\
OFA & 0.16 & 0.59 & 0.24 & 0.71 & 0.22 & 0.67 & 0.45 & 0.20 & 0.86 \\
Sub-MCD & 0.15 & 0.67 & 0.24 & 0.72 & 0.23 & 0.54 & 0.32 & 0.24 & 0.81 \\
Sub-HBOS & 0.18 & 0.61 & 0.23 & 0.67 & 0.23 & 0.60 & 0.35 & 0.27 & 0.79 \\
Sub-OCSVM & 0.16 & 0.65 & 0.23 & 0.73 & 0.22 & 0.55 & 0.32 & 0.23 & 0.79 \\
Sub-IForest & 0.16 & 0.63 & 0.22 & 0.72 & 0.22 & 0.63 & 0.34 & 0.23 & 0.80 \\
Donut & 0.14 & 0.56 & 0.20 & 0.68 & 0.20 & 0.57 & 0.38 & 0.20 & 0.82 \\
LOF & 0.14 & 0.58 & 0.17 & 0.68 & 0.21 & 0.63 & 0.40 & 0.22 & 0.79 \\
AnomalyTransformer & 0.08 & 0.50 & 0.12 & 0.56 & 0.12 & 0.53 & 0.34 & 0.14 & 0.77 \\
\bottomrule
\end{tabular}
\end{table}

\begin{table}[!htp]
\centering
\caption{Results on TSB-AD-M (180 evaluation sequences).}
\label{tab:updated-m-metrics}
\AppendixTableStyle
\setlength{\tabcolsep}{3.0pt}
\begin{tabular}{lrrrrrrrrr}
\toprule
Method & AUC-PR & AUC-ROC & VUS-PR & VUS-ROC & Std.-F1 & PA-F1 & Event-F1 & R-F1 & Aff.-F \\
\midrule
HieraNTP & \textbf{0.514} & \textbf{0.865} & \textbf{0.503} & \textbf{0.870} & -- & -- & -- & -- & -- \\
TSPulse (FT) & -- & -- & 0.387 & -- & -- & -- & -- & -- & -- \\
TSPulse (ZS) & -- & -- & 0.361 & -- & -- & -- & -- & -- & -- \\
xLSTMAD & 0.344 & 0.739 & 0.392 & 0.778 & 0.425 & 0.621 & 0.532 & 0.390 & 0.833 \\
MMPAD & 0.309 & 0.744 & 0.354 & 0.759 & 0.359 & 0.480 & 0.391 & 0.300 & 0.814 \\
CHARM & -- & -- & 0.391 & -- & -- & -- & -- & -- & -- \\
Time-RCD & 0.340 & 0.722 & 0.317 & 0.707 & 0.384 & 0.681 & -- & -- & 0.823 \\
StreamVAE & -- & -- & 0.435 & -- & -- & -- & -- & -- & -- \\
CNN & 0.32 & 0.73 & 0.31 & 0.76 & 0.37 & 0.78 & 0.65 & 0.37 & 0.87 \\
OmniAnomaly & 0.27 & 0.65 & 0.31 & 0.69 & 0.32 & 0.55 & 0.41 & 0.37 & 0.81 \\
PCA & 0.31 & 0.70 & 0.31 & 0.74 & 0.37 & 0.79 & 0.59 & 0.29 & 0.85 \\
LSTMAD & 0.31 & 0.70 & 0.31 & 0.74 & 0.36 & 0.79 & 0.64 & 0.38 & 0.87 \\
USAD & 0.26 & 0.64 & 0.30 & 0.68 & 0.31 & 0.53 & 0.40 & 0.37 & 0.80 \\
AutoEncoder & 0.30 & 0.67 & 0.30 & 0.69 & 0.34 & 0.60 & 0.44 & 0.28 & 0.80 \\
KMeansAD & 0.25 & 0.69 & 0.29 & 0.73 & 0.31 & 0.68 & 0.49 & 0.33 & 0.82 \\
CBLOF & 0.28 & 0.67 & 0.27 & 0.70 & 0.32 & 0.65 & 0.45 & 0.31 & 0.81 \\
MCD & 0.27 & 0.65 & 0.27 & 0.69 & 0.33 & 0.46 & 0.33 & 0.20 & 0.76 \\
OCSVM & 0.23 & 0.61 & 0.26 & 0.67 & 0.28 & 0.48 & 0.41 & 0.30 & 0.80 \\
Donut & 0.20 & 0.64 & 0.26 & 0.71 & 0.28 & 0.52 & 0.36 & 0.21 & 0.81 \\
RobustPCA & 0.24 & 0.58 & 0.24 & 0.61 & 0.29 & 0.60 & 0.42 & 0.33 & 0.81 \\
FITS & 0.15 & 0.58 & 0.21 & 0.66 & 0.22 & 0.72 & 0.32 & 0.16 & 0.81 \\
OFA & 0.15 & 0.55 & 0.21 & 0.63 & 0.21 & 0.72 & 0.41 & 0.17 & 0.83 \\
EIF & 0.19 & 0.67 & 0.21 & 0.71 & 0.26 & 0.74 & 0.44 & 0.26 & 0.81 \\
COPOD & 0.20 & 0.65 & 0.20 & 0.69 & 0.27 & 0.72 & 0.41 & 0.24 & 0.80 \\
IForest & 0.19 & 0.66 & 0.20 & 0.69 & 0.26 & 0.68 & 0.41 & 0.24 & 0.80 \\
HBOS & 0.16 & 0.63 & 0.19 & 0.67 & 0.24 & 0.67 & 0.40 & 0.24 & 0.80 \\
TimesNet & 0.13 & 0.56 & 0.19 & 0.64 & 0.20 & 0.68 & 0.32 & 0.17 & 0.82 \\
KNN & 0.14 & 0.51 & 0.18 & 0.59 & 0.19 & 0.69 & 0.45 & 0.21 & 0.79 \\
TranAD & 0.14 & 0.59 & 0.18 & 0.65 & 0.21 & 0.68 & 0.40 & 0.21 & 0.79 \\
LOF & 0.10 & 0.53 & 0.14 & 0.60 & 0.15 & 0.57 & 0.32 & 0.14 & 0.76 \\
AnomalyTransformer & 0.07 & 0.52 & 0.12 & 0.57 & 0.12 & 0.53 & 0.33 & 0.14 & 0.74 \\
\bottomrule
\end{tabular}
\end{table}

TSPulse, CHARM, and StreamVAE publish only VUS-PR for the
corresponding submissions.

\subsection{Per-Dataset TSB-AD Results}
\label{app:per-dataset}

Tables~\ref{tab:u-per-dataset} and~\ref{tab:m-per-dataset} report HieraNTP results for all datasets. Each dataset row is the
mean over its Eval sequences; the overall row averages all Eval sequences.
AP denotes average precision.

\begin{table}[!htp]
\centering
\caption{Per-dataset results on the 350 TSB-AD-U evaluation sequences.}
\label{tab:u-per-dataset}
\AppendixTableStyle
\setlength{\tabcolsep}{7pt}
\begin{tabular}{lccccc}
\toprule
Dataset & Sequences & AP & AUC-ROC & VUS-PR & VUS-ROC \\
\midrule
CATSv2 & 1 & 0.503 & 0.725 & 0.375 & 0.741 \\
Daphnet & 1 & 0.176 & 0.871 & 0.188 & 0.876 \\
Exathlon & 30 & 0.950 & 0.996 & 0.950 & 0.996 \\
IOPS & 15 & 0.477 & 0.958 & 0.515 & 0.977 \\
LTDB & 8 & 0.329 & 0.619 & 0.408 & 0.711 \\
MGAB & 8 & 0.004 & 0.612 & 0.010 & 0.686 \\
MITDB & 7 & 0.289 & 0.742 & 0.316 & 0.789 \\
MSL & 7 & 0.666 & 0.957 & 0.777 & 0.975 \\
NAB & 23 & 0.735 & 0.932 & 0.752 & 0.937 \\
NEK & 8 & 0.833 & 0.960 & 0.853 & 0.975 \\
OPPORTUNITY & 27 & 0.811 & 0.962 & 0.808 & 0.963 \\
Power & 1 & 0.109 & 0.600 & 0.110 & 0.608 \\
SED & 2 & 0.645 & 0.976 & 0.849 & 0.992 \\
SMAP & 17 & 0.865 & 0.991 & 0.899 & 0.995 \\
SMD & 33 & 0.819 & 0.979 & 0.826 & 0.980 \\
Stock & 8 & 0.980 & 0.999 & 0.982 & 0.999 \\
SVDB & 18 & 0.539 & 0.901 & 0.518 & 0.927 \\
SWaT & 1 & 0.714 & 0.835 & 0.626 & 0.798 \\
TAO & 2 & 0.999 & 1.000 & 1.000 & 1.000 \\
TODS & 13 & 0.629 & 0.901 & 0.815 & 0.944 \\
UCR & 70 & 0.337 & 0.873 & 0.349 & 0.889 \\
WSD & 20 & 0.690 & 0.991 & 0.643 & 0.993 \\
YAHOO & 30 & 0.826 & 0.991 & 0.790 & 0.991 \\
\midrule
Overall & 350 & 0.643 & 0.925 & 0.656 & 0.939 \\
\bottomrule
\end{tabular}
\end{table}

\begin{table}[!htp]
\centering
\caption{Per-dataset results on the 180 TSB-AD-M evaluation sequences.}
\label{tab:m-per-dataset}
\AppendixTableStyle
\setlength{\tabcolsep}{7pt}
\begin{tabular}{lccccc}
\toprule
Dataset & Sequences & AP & AUC-ROC & VUS-PR & VUS-ROC \\
\midrule
CATSv2 & 5 & 0.256 & 0.796 & 0.221 & 0.799 \\
CreditCard & 1 & 0.041 & 0.758 & 0.062 & 0.753 \\
Daphnet & 1 & 0.387 & 0.938 & 0.400 & 0.941 \\
Exathlon & 25 & 0.999 & 1.000 & 0.998 & 1.000 \\
GECCO & 1 & 0.346 & 0.789 & 0.099 & 0.675 \\
Genesis & 1 & 0.005 & 0.724 & 0.015 & 0.837 \\
GHL & 23 & 0.112 & 0.692 & 0.098 & 0.700 \\
LTDB & 4 & 0.564 & 0.858 & 0.583 & 0.877 \\
MITDB & 11 & 0.341 & 0.818 & 0.267 & 0.839 \\
MSL & 14 & 0.301 & 0.739 & 0.340 & 0.764 \\
OPPORTUNITY & 7 & 0.183 & 0.671 & 0.172 & 0.666 \\
PSM & 1 & 0.148 & 0.639 & 0.174 & 0.612 \\
SMAP & 25 & 0.484 & 0.913 & 0.524 & 0.922 \\
SMD & 20 & 0.495 & 0.889 & 0.392 & 0.861 \\
SVDB & 28 & 0.616 & 0.936 & 0.623 & 0.955 \\
SWaT & 2 & 0.544 & 0.832 & 0.432 & 0.777 \\
TAO & 11 & 0.991 & 0.999 & 0.995 & 0.999 \\
\midrule
Overall & 180 & 0.514 & 0.865 & 0.503 & 0.870 \\
\bottomrule
\end{tabular}
\end{table}

\subsection{TAB Full-Shot Evaluation and OpenTS Comparison}
\label{app:opents-protocol}

We use TAB's full-shot evaluation strategy for univariate and multivariate
anomaly detection \citep{qiu2025tab}. HieraNTP fits its task-adaptive detector
with labels from the official training prefixes and retains the frozen
pretrained tokenizer and predictor. Training pools all 15 univariate or 29
multivariate datasets into a shared base detector per track, followed by
dataset-specific extensions. The split follows
the training and test lengths in TAB's metadata. Detector fitting uses sampled
training-prefix points, and evaluation uses the test suffixes. Sampling, weights,
and boosting parameters are given in Appendix~\ref{app:tab-detector-training}.
The OpenTS leaderboard contains baseline
results for 14 of TAB's 15 univariate datasets and 28 of its 29
multivariate datasets. OPP, named OPPORTUNITY in the local data files, has
no baseline entry and is excluded from the comparison in
Table~\ref{tab:opents-main}. API names \texttt{Creditcard},
\texttt{M\_genesis}, and \texttt{KDD-99} map to Credit, Genesis, and KDDcup99.

TAB reports ten label-based and six score-based metrics. The OpenTS fields
\texttt{A-P}, \texttt{A-R}, and \texttt{V-PR} denote AUC-PR, AUC-ROC,
and VUS-PR, respectively. TAB computes AUC-PR with average precision.
The metric columns in Table~\ref{tab:opents-main} are arithmetic means of
the dataset-level public values on the common 14/28-dataset sets. Public
API values have three decimal places, and HieraNTP results are rounded to the
same precision before computing those means. TSB-AD averages its Eval
sequences, while this TAB comparison averages datasets.

HieraNTP evaluates label-based metrics at anomaly ratios of 0.1\%, 0.5\%,
1\%, 2\%, 3\%, 5\%, 10\%, 15\%, 20\%, and 25\%. Each ratio defines a
candidate threshold for each test series. The cutoff is the corresponding
percentile of the combined
training and test anomaly scores, without using labels to set the cutoff.
Evaluation labels select the ratio maximizing Affiliation-F1, and the other
label-based metrics are read at that threshold. TAB describes reporting the
best result across candidate thresholds. Its threshold selection differs from
the shared Affiliation-F1 rule used for HieraNTP. The F1-based columns
describe each method under its stated evaluation rule. AP, AUC-ROC, VUS-PR,
and VUS-ROC are computed from the original anomaly scores.

Medals are recomputed after adding HieraNTP to each public dataset--metric
cell. The three highest distinct values receive gold, silver, and bronze
($G/S/B$), with ties sharing medals. All 16 metrics contribute to the weighted
composite score $3G+2S+B$.

% Source: TAB25/artifacts/tab_mts_indomain/opents_base384_family32_recomputed_20260924.
% Use OURS_FAMILY32 only; public baseline values and per-dataset values retain 3dp.
\FloatBarrier
\subsection{Extended TAB Model Comparisons}
\label{app:tab-model-results}
Tables~\ref{tab:tab-all-u} and~\ref{tab:tab-all-m} compare HieraNTP with
OpenTS methods reporting at least one of the six metrics.
Means use the common 14/28-dataset comparison sets, following
Appendix~\ref{app:opents-protocol}.
Bold entries mark the highest dataset-mean value in each column.

\begin{table}[!htp]
\centering
\caption{Extended TAB univariate comparison on 14 datasets.}
\label{tab:tab-all-u}
\AppendixTableStyle
\setlength{\tabcolsep}{5pt}
\begin{tabular}{@{}lcccccc@{}}
\toprule
Method & F1 & Aff-F1 & AP & AUC-ROC & VUS-PR & VUS-ROC \\
\midrule
\rowcolor{gray!12} HieraNTP & \textbf{0.322} & \textbf{0.849} & \textbf{0.419} & \textbf{0.817} & \textbf{0.439} & \textbf{0.837} \\
KMeans & 0.293 & 0.792 & 0.356 & 0.790 & 0.362 & 0.784 \\
DWT-MLEAD & 0.242 & 0.716 & 0.291 & 0.739 & 0.312 & 0.699 \\
CALF & 0.182 & 0.820 & 0.231 & 0.687 & 0.251 & 0.758 \\
iTransformer & 0.194 & 0.810 & 0.243 & 0.687 & 0.244 & 0.753 \\
HBOS & 0.107 & 0.696 & 0.104 & 0.599 & 0.242 & 0.625 \\
TimesNet & 0.172 & 0.805 & 0.224 & 0.677 & 0.224 & 0.742 \\
UniTS & 0.161 & 0.798 & 0.208 & 0.651 & 0.221 & 0.718 \\
OCSVM & 0.107 & 0.700 & 0.148 & 0.619 & 0.218 & 0.658 \\
TimesFM & 0.196 & 0.808 & 0.223 & 0.622 & 0.217 & 0.677 \\
GPT4TS & 0.173 & 0.805 & 0.223 & 0.667 & 0.215 & 0.724 \\
KNN & 0.109 & 0.708 & 0.192 & 0.626 & 0.213 & 0.638 \\
ModernTCN & 0.163 & 0.807 & 0.210 & 0.663 & 0.210 & 0.725 \\
UniTime & 0.163 & 0.803 & 0.204 & 0.662 & 0.209 & 0.729 \\
PatchTST & 0.157 & 0.803 & 0.208 & 0.667 & 0.208 & 0.728 \\
DAGMM & 0.023 & 0.571 & 0.104 & 0.548 & 0.204 & 0.599 \\
DLinear & 0.165 & 0.808 & 0.213 & 0.659 & 0.202 & 0.721 \\
LLMMixer & 0.148 & 0.804 & 0.196 & 0.654 & 0.202 & 0.723 \\
EIF & 0.166 & 0.782 & 0.213 & 0.633 & 0.202 & 0.671 \\
LODA & 0.101 & 0.674 & 0.104 & 0.576 & 0.197 & 0.608 \\
Timer & 0.157 & 0.774 & 0.198 & 0.615 & 0.195 & 0.683 \\
Chronos & 0.160 & 0.796 & 0.191 & 0.629 & 0.195 & 0.682 \\
LSTMED & 0.079 & 0.704 & 0.213 & 0.619 & 0.193 & 0.669 \\
Moment & 0.160 & 0.796 & 0.189 & 0.626 & 0.191 & 0.695 \\
AE & 0.076 & 0.682 & 0.198 & 0.574 & 0.189 & 0.609 \\
ZMS & 0.073 & 0.709 & 0.169 & 0.592 & 0.184 & 0.650 \\
NLinear & 0.144 & 0.794 & 0.171 & 0.653 & 0.183 & 0.719 \\
TranAD & 0.158 & 0.774 & 0.201 & 0.545 & 0.179 & 0.594 \\
VAE & 0.071 & 0.681 & 0.200 & 0.574 & 0.177 & 0.622 \\
Torsk & 0.141 & 0.703 & 0.143 & 0.681 & 0.177 & 0.662 \\
Dada & 0.137 & 0.801 & 0.187 & 0.625 & 0.174 & 0.682 \\
CBLOF & 0.100 & 0.692 & 0.189 & 0.603 & 0.173 & 0.651 \\
PCA & 0.095 & 0.674 & 0.188 & 0.573 & 0.172 & 0.606 \\
LOF & 0.098 & 0.706 & 0.149 & 0.613 & 0.169 & 0.638 \\
IF & 0.089 & 0.718 & 0.132 & 0.604 & 0.168 & 0.662 \\
TTM & 0.120 & 0.755 & 0.170 & 0.600 & 0.167 & 0.663 \\
StatThreshold & 0.065 & 0.637 & 0.140 & 0.528 & 0.148 & 0.578 \\
Series2Graph & 0.114 & 0.781 & 0.090 & 0.633 & 0.129 & 0.675 \\
ARIMA & 0.070 & 0.652 & 0.117 & 0.508 & 0.129 & 0.564 \\
SARIMA & 0.071 & 0.682 & 0.114 & 0.525 & 0.127 & 0.586 \\
SpectralResidual & 0.065 & 0.698 & 0.114 & 0.533 & 0.123 & 0.610 \\
DualTF & 0.078 & 0.728 & 0.083 & 0.539 & 0.108 & 0.604 \\
LeftSTAMPi & 0.084 & 0.726 & 0.066 & 0.534 & 0.104 & 0.583 \\
ContraAD & 0.041 & 0.684 & 0.033 & 0.499 & 0.059 & 0.579 \\
ATrans & 0.025 & 0.700 & 0.038 & 0.473 & 0.058 & 0.485 \\
DCdetector & 0.024 & 0.688 & 0.034 & 0.473 & 0.057 & 0.491 \\
\bottomrule
\end{tabular}
\end{table}

\begin{table}[!htp]
\centering
\caption{Extended TAB multivariate comparison on 28 datasets.}
\label{tab:tab-all-m}
\AppendixTableStyle
\setlength{\tabcolsep}{5pt}
\begin{tabular}{@{}lcccccc@{}}
\toprule
Method & F1 & Aff-F1 & AP & AUC-ROC & VUS-PR & VUS-ROC \\
\midrule
\rowcolor{gray!12} HieraNTP & \textbf{0.331} & \textbf{0.793} & \textbf{0.376} & \textbf{0.782} & \textbf{0.388} & \textbf{0.794} \\
OCSVM & 0.161 & 0.677 & 0.176 & 0.645 & 0.309 & 0.656 \\
KMeans & 0.296 & 0.698 & 0.340 & 0.744 & 0.302 & 0.685 \\
KNN & 0.179 & 0.699 & 0.336 & 0.714 & 0.301 & 0.707 \\
DAGMM & 0.083 & 0.561 & 0.171 & 0.648 & 0.294 & 0.660 \\
TimesNet & 0.199 & 0.789 & 0.259 & 0.688 & 0.271 & 0.722 \\
LSTMED & 0.109 & 0.587 & 0.281 & 0.686 & 0.269 & 0.696 \\
VAE & 0.103 & 0.554 & 0.278 & 0.689 & 0.263 & 0.694 \\
CATCH & 0.191 & 0.778 & 0.240 & 0.668 & 0.261 & 0.705 \\
TranAD & 0.238 & 0.759 & 0.280 & 0.685 & 0.259 & 0.680 \\
AE & 0.109 & 0.621 & 0.285 & 0.697 & 0.258 & 0.693 \\
CBLOF & 0.185 & 0.676 & 0.273 & 0.710 & 0.258 & 0.716 \\
EIF & 0.210 & 0.773 & 0.254 & 0.732 & 0.258 & 0.741 \\
DUET & 0.200 & 0.780 & 0.239 & 0.663 & 0.258 & 0.699 \\
PatchTST & 0.196 & 0.785 & 0.240 & 0.677 & 0.257 & 0.716 \\
iTransformer & 0.183 & 0.770 & 0.235 & 0.667 & 0.251 & 0.705 \\
HBOS & 0.170 & 0.660 & 0.250 & 0.664 & 0.245 & 0.677 \\
ModernTCN & 0.187 & 0.776 & 0.229 & 0.656 & 0.244 & 0.695 \\
DLinear & 0.176 & 0.775 & 0.227 & 0.660 & 0.240 & 0.695 \\
LOF & 0.164 & 0.693 & 0.245 & 0.635 & 0.236 & 0.631 \\
NLinear & 0.170 & 0.773 & 0.211 & 0.665 & 0.232 & 0.701 \\
PCA & 0.155 & 0.651 & 0.244 & 0.686 & 0.229 & 0.672 \\
IF & 0.078 & 0.579 & 0.199 & 0.636 & 0.212 & 0.675 \\
LODA & 0.157 & 0.656 & 0.187 & 0.578 & 0.181 & 0.604 \\
DualTF & 0.109 & 0.593 & 0.147 & 0.611 & 0.174 & 0.621 \\
Torsk & 0.139 & 0.602 & 0.137 & 0.575 & 0.171 & 0.555 \\
ContraAD & 0.103 & 0.621 & 0.118 & 0.498 & 0.137 & 0.516 \\
ATrans & 0.055 & 0.653 & 0.092 & 0.500 & 0.122 & 0.499 \\
DCdetector & 0.046 & 0.637 & 0.085 & 0.496 & 0.118 & 0.503 \\
\bottomrule
\end{tabular}
\end{table}

\FloatBarrier
\subsection{Complete Per-Dataset TAB Results}
\label{app:tab-per-dataset}
Tables~\ref{tab:tab-datasets-u} and~\ref{tab:tab-datasets-m} report
HieraNTP on all 15 univariate and 29 multivariate TAB datasets, including
OPP (OPPORTUNITY). Dataset names follow OpenTS.
The comparison mean excludes OPP, matching Table~\ref{tab:opents-main};
the all-dataset mean includes it. Values retain three decimals.

\begin{table}[!htp]
\centering
\caption{HieraNTP results on all 15 TAB univariate datasets.}
\label{tab:tab-datasets-u}
\AppendixTableStyle
\setlength{\tabcolsep}{5pt}
\begin{tabular}{@{}lcccccc@{}}
\toprule
Dataset & F1 & Aff-F1 & AP & AUC-ROC & VUS-PR & VUS-ROC \\
\midrule
Daphnet & 0.051 & 0.712 & 0.063 & 0.642 & 0.112 & 0.715 \\
ECG & 0.616 & 0.903 & 0.728 & 0.969 & 0.842 & 0.984 \\
GAIA & 0.262 & 0.877 & 0.444 & 0.906 & 0.382 & 0.909 \\
Genesis & 0.523 & 0.974 & 0.630 & 0.983 & 0.740 & 0.994 \\
GHL & 0.000 & 0.714 & 0.003 & 0.188 & 0.005 & 0.305 \\
IOPS & 0.336 & 0.898 & 0.462 & 0.923 & 0.478 & 0.934 \\
KDD21 & 0.152 & 0.833 & 0.226 & 0.854 & 0.244 & 0.871 \\
MGAB & 0.009 & 0.673 & 0.004 & 0.585 & 0.006 & 0.590 \\
MSL & 0.359 & 0.868 & 0.446 & 0.854 & 0.518 & 0.881 \\
NAB & 0.357 & 0.915 & 0.546 & 0.874 & 0.611 & 0.885 \\
OPPORTUNITY & 0.527 & 0.952 & 0.706 & 0.965 & 0.624 & 0.950 \\
SMAP & 0.366 & 0.905 & 0.574 & 0.931 & 0.551 & 0.923 \\
SMD & 0.318 & 0.813 & 0.385 & 0.851 & 0.360 & 0.829 \\
SVDB & 0.471 & 0.862 & 0.509 & 0.905 & 0.554 & 0.933 \\
YAHOO & 0.683 & 0.942 & 0.852 & 0.977 & 0.749 & 0.967 \\
\midrule
Comparison mean & 0.322 & 0.849 & 0.419 & 0.817 & 0.439 & 0.837 \\
All-dataset mean & 0.335 & 0.856 & 0.439 & 0.827 & 0.452 & 0.845 \\
\bottomrule
\end{tabular}
\end{table}

\begin{table}[!htp]
\centering
\caption{HieraNTP results on all 29 TAB multivariate datasets.}
\label{tab:tab-datasets-m}
\AppendixTableStyle
\setlength{\tabcolsep}{5pt}
\begin{tabular}{@{}lcccccc@{}}
\toprule
Dataset & F1 & Aff-F1 & AP & AUC-ROC & VUS-PR & VUS-ROC \\
\midrule
ASD & 0.384 & 0.832 & 0.377 & 0.837 & 0.389 & 0.862 \\
CalIt2 & 0.149 & 0.780 & 0.076 & 0.811 & 0.102 & 0.807 \\
CATSv2 & 0.134 & 0.689 & 0.117 & 0.663 & 0.147 & 0.709 \\
CICIDS & 0.040 & 0.854 & 0.125 & 0.958 & 0.153 & 0.931 \\
Creditcard & 0.078 & 0.754 & 0.128 & 0.955 & 0.098 & 0.914 \\
DAPHNET & 0.277 & 0.828 & 0.262 & 0.787 & 0.372 & 0.843 \\
DLR & 0.808 & 0.999 & 0.922 & 0.999 & 0.964 & 0.999 \\
ECG & 0.482 & 0.769 & 0.509 & 0.839 & 0.458 & 0.801 \\
Exathlon & 0.850 & 0.999 & 0.998 & 1.000 & 0.958 & 0.986 \\
GECCO & 0.269 & 0.917 & 0.530 & 0.976 & 0.455 & 0.983 \\
GHL & 0.013 & 0.596 & 0.018 & 0.396 & 0.027 & 0.471 \\
GutenTAG & 0.534 & 0.951 & 0.712 & 0.935 & 0.814 & 0.944 \\
KDD-99 & 0.198 & 0.758 & 0.121 & 0.973 & 0.102 & 0.964 \\
LTDB & 0.657 & 0.910 & 0.733 & 0.946 & 0.835 & 0.970 \\
M\_genesis & 0.003 & 0.675 & 0.003 & 0.442 & 0.008 & 0.609 \\
MITDB & 0.213 & 0.865 & 0.305 & 0.857 & 0.304 & 0.890 \\
MSL & 0.088 & 0.657 & 0.079 & 0.381 & 0.129 & 0.496 \\
NYC & 0.075 & 0.541 & 0.035 & 0.701 & 0.049 & 0.711 \\
OPPORTUNITY & 0.132 & 0.777 & 0.119 & 0.929 & 0.141 & 0.934 \\
PSM & 0.344 & 0.783 & 0.424 & 0.650 & 0.418 & 0.636 \\
PUMP & 0.334 & 0.704 & 0.161 & 0.733 & 0.198 & 0.733 \\
SKAB & 0.428 & 0.827 & 0.480 & 0.563 & 0.595 & 0.633 \\
SMAP & 0.288 & 0.674 & 0.221 & 0.633 & 0.239 & 0.653 \\
SMD & 0.129 & 0.669 & 0.079 & 0.661 & 0.079 & 0.638 \\
SVDB & 0.541 & 0.939 & 0.675 & 0.972 & 0.687 & 0.978 \\
SWAN & 0.363 & 0.746 & 0.460 & 0.622 & 0.452 & 0.576 \\
SWAT & 0.403 & 0.758 & 0.731 & 0.858 & 0.624 & 0.833 \\
TAO & 0.977 & 0.993 & 0.999 & 1.000 & 0.951 & 0.985 \\
TODS & 0.215 & 0.742 & 0.258 & 0.738 & 0.249 & 0.682 \\
\midrule
Comparison mean & 0.331 & 0.793 & 0.376 & 0.782 & 0.388 & 0.794 \\
All-dataset mean & 0.324 & 0.793 & 0.367 & 0.787 & 0.379 & 0.799 \\
\bottomrule
\end{tabular}
\end{table}

\FloatBarrier

\section{Ablation Studies}
\label{app:additional-ablations}

\begin{table}[!tp]
\centering
\caption{Complete TSB-AD-U ablations: sequence-mean metrics (\%, higher is better).}
\label{tab:full-ablation}
\AppendixTableStyle
\setlength{\tabcolsep}{5pt}
\begin{tabular}{@{}lcccc@{}}
\toprule
Configuration & VUS-PR & AP & VUS-ROC & AUC-ROC \\
\midrule
HieraNTP & 65.56 & 64.26 & 93.85 & 92.51 \\
Shared base detector only & 61.48 & 59.93 & 93.36 & 91.87 \\
\midrule
Time-domain error & 33.85 & 34.42 & 80.36 & 76.18 \\
Fixed evidence averaging & 43.65 & 39.96 & 87.42 & 84.61 \\
Time-domain error + detector & 29.89 & 29.71 & 78.77 & 73.77 \\
Mean + detector & 38.20 & 32.71 & 86.72 & 82.56 \\
Max + detector & 32.11 & 25.91 & 76.33 & 74.47 \\
Mean--max + detector & 44.51 & 38.13 & 86.87 & 83.77 \\
Instantaneous multiview & 57.80 & 56.78 & 92.27 & 89.85 \\
Prediction-to-observation only & 59.78 & 58.23 & 90.96 & 89.12 \\
\midrule
w/o reference calibration & 60.28 & 58.75 & 91.53 & 89.60 \\
w/o temporal summaries & 59.25 & 58.46 & 92.61 & 90.36 \\
Forward only & 60.32 & 58.93 & 92.73 & 91.17 \\
\midrule
w/o token-space & 65.35 & 63.98 & 93.94 & 92.52 \\
w/o patch mean & 64.31 & 62.71 & 93.50 & 92.07 \\
w/o patch variance & 62.47 & 61.35 & 92.06 & 90.49 \\
w/o observation-space & 62.54 & 56.84 & 92.72 & 90.71 \\
w/o full-resolution decoded discrepancies & 63.23 & 57.94 & 92.91 & 91.00 \\
w/o native-scale decoded discrepancies & 65.19 & 64.05 & 93.91 & 92.48 \\
w/o decoded observation & 64.67 & 62.40 & 93.69 & 92.29 \\
w/o decoded target & 65.88 & 64.56 & 93.83 & 92.50 \\
\midrule
w/o patch 256 & 65.53 & 64.59 & 93.58 & 92.12 \\
w/o patch 128 & 65.52 & 64.17 & 93.83 & 92.46 \\
w/o patch 64 & 65.83 & 64.53 & 93.75 & 92.33 \\
w/o patch 32 & 66.36 & 65.16 & 93.93 & 92.63 \\
w/o patch 16 & 66.12 & 64.61 & 93.89 & 92.53 \\
w/o patch 8 & 63.46 & 62.24 & 93.86 & 92.41 \\
Single token-grid offset & 64.22 & 62.38 & 93.50 & 92.12 \\
\bottomrule
\end{tabular}
\end{table}

\begin{table}[!tp]
\centering
\caption{Complete TSB-AD-M ablations: sequence-mean metrics (\%, higher is better).}
\label{tab:full-ablation-m}
\AppendixTableStyle
\setlength{\tabcolsep}{5pt}
\begin{tabular}{@{}lcccc@{}}
\toprule
Configuration & VUS-PR & AP & VUS-ROC & AUC-ROC \\
\midrule
HieraNTP & 50.28 & 51.39 & 86.99 & 86.48 \\
Shared base detector only & 49.91 & 51.04 & 86.79 & 86.33 \\
\midrule
Time-domain error & 30.81 & 30.91 & 74.86 & 72.06 \\
Fixed evidence averaging & 35.27 & 34.09 & 77.53 & 75.74 \\
Time-domain error + detector & 28.34 & 28.44 & 74.09 & 70.57 \\
Mean + detector & 29.12 & 24.77 & 78.22 & 74.25 \\
Max + detector & 27.47 & 22.32 & 67.65 & 66.18 \\
Mean--max + detector & 33.83 & 29.66 & 76.29 & 73.33 \\
Instantaneous multiview & 44.08 & 44.38 & 84.01 & 82.34 \\
Prediction-to-observation only & 44.60 & 45.59 & 84.39 & 83.47 \\
\midrule
w/o reference calibration & 51.71 & 52.73 & 88.09 & 87.18 \\
w/o temporal summaries & 45.63 & 46.16 & 85.58 & 84.31 \\
Forward only & 47.52 & 48.53 & 86.07 & 85.35 \\
Variable mean only & 47.55 & 48.13 & 85.56 & 84.85 \\
Variable maximum only & 47.41 & 47.61 & 85.31 & 84.64 \\
\midrule
w/o token-space & 49.69 & 50.31 & 86.52 & 85.77 \\
w/o patch mean & 50.33 & 51.23 & 87.06 & 86.47 \\
w/o patch variance & 47.25 & 48.81 & 85.83 & 85.27 \\
w/o observation-space & 47.77 & 44.67 & 86.05 & 84.50 \\
w/o full-resolution decoded discrepancies & 48.03 & 45.35 & 86.16 & 84.55 \\
w/o native-scale decoded discrepancies & 49.57 & 50.53 & 87.31 & 86.85 \\
w/o decoded observation & 49.14 & 49.61 & 87.02 & 86.32 \\
w/o decoded target & 49.51 & 50.36 & 86.71 & 86.24 \\
\midrule
w/o patch 256 & 49.98 & 51.06 & 86.42 & 85.86 \\
w/o patch 128 & 49.29 & 50.56 & 87.16 & 86.48 \\
w/o patch 64 & 49.92 & 50.86 & 86.83 & 86.22 \\
w/o patch 32 & 50.08 & 50.98 & 86.79 & 86.06 \\
w/o patch 16 & 49.70 & 50.78 & 86.93 & 86.24 \\
w/o patch 8 & 49.46 & 50.59 & 86.99 & 86.36 \\
Single token-grid offset & 49.62 & 50.46 & 87.27 & 86.66 \\
\bottomrule
\end{tabular}
\end{table}

Tables~\ref{tab:full-ablation} and~\ref{tab:full-ablation-m} report VUS-PR,
average precision (AP), VUS-ROC, and AUC-ROC for the representation and
adaptation comparisons. All configurations evaluate the same 350 U or 180 M
sequences. All detector-based comparisons use the original training samples,
labels, weights, random seed, and hyperparameters. When the detector input
changes, the shared base detector and applicable dataset-specific extensions
are refitted under the same evaluation protocol. Scores average both directions
except in the forward-only configuration.
Dataset routing is unchanged except for Shared base detector only.
The fixed scorers defined
below use no fitted trees. Component-removal variants modify the final
2496D/4992D representation.
The patch-length rows remove the corresponding discrepancy channels and derived
summaries while retaining all tokenizer and predictor layers.
Full-resolution decoded discrepancies remain in those rows and still depend on
the complete token hierarchy. The reference-calibration ablation retains the
reference-derived precision matrices, support masks, and fixed clipping rules. The
forward-only configuration retains centered temporal summaries. All results
use fixed single-run configurations.

\paragraph{Aggregation before supervised detection.}
\label{app:supervised-aggregation}
% Author-supplied completed export: data/aggregation_supervised_results.csv.
Mean + detector and Max + detector reduce each complete 2496D or 4992D
feature vector to its arithmetic mean or maximum, respectively.
Mean--max + detector concatenates both scalars. Aggregation operates across the final feature
coordinates at each timestamp, using their existing values without additional
cross-feature standardization.

\paragraph{Prediction-to-observation evidence.}
\label{app:prediction-observation-control}
% Source: data/prediction_observation_results.csv; T10-B in the P1 materials.
Prediction-to-observation only retains the decoded-observation source,
comparing decoded predictions with observations at six native scales and full resolution.
The two temporal alignments yield 14 base channels. Reference calibration,
the three processing variants, and all 13 temporal summaries are unchanged,
giving 546 features on U and 1092 on M with both variable means and maxima.
Columns are selected directly from the complete feature matrices in their
original relative order. This comparison evaluates complementary evidence
sources with the tokenizer, predictor, and decoder fixed.

\paragraph{Stage-wise evidence construction.}
\label{app:stagewise-evidence}
% Full-precision results and export paths: data/p0_stage_results.csv.
Time-domain error uses the scalar time-domain prediction error
directly as the anomaly score. Time-domain error + detector
supplies the same scalar to the supervised downstream
detector. Instantaneous multiview evidence replaces this scalar with 64
instantaneous features on U, or 128 after variable mean/max aggregation on M.
The next stage incorporates all three
processing variants while retaining only their current values, corresponding
to w/o temporal summaries. HieraNTP adds the temporal summaries
to form the complete representation. Shared base detector only evaluates this
representation without dataset-specific extensions. Fixed evidence averaging is a distinct label-free reference,
defined below.

\paragraph{Fixed evidence averaging.}
\label{app:fixed-averaging}
% Sources: HieraNTP64/results/zero_shot_four_metrics.json;
% nogeo64_ablation_export_v2/previous_export/code_snapshot/run_parentfree64_ablation.py.
The fixed scorer uses the pool-then-calibrate evidence before temporal
summaries. It takes the positive part of each channel, averages over the 64
univariate or 128 multivariate channels, and averages the forward and reverse
scores. It uses neither anomaly labels nor fitted trees and retains the same
64 base discrepancy definitions as the complete method. Its comparison with
HieraNTP changes both the representation and the scoring rule.

\paragraph{Target-dataset label availability.}
\label{app:label-availability}
We partition the existing full-model Eval results according to whether the
target dataset contributes anomaly-labeled samples to the detector's Tuning
set. Every dataset represented in that training set contains sampled anomalous
points. The remaining U datasets are CATSv2, Daphnet, Power, and SWaT;
the remaining M datasets are CreditCard, Daphnet, GECCO, Genesis, PSM, and
SWaT. These datasets use the shared supervised detector trained on other
datasets. This grouping concerns the detector's actual training samples,
independently of whether an Eval sequence has a normal training prefix.
Table~\ref{tab:label-availability} reports sequence-mean metrics for each
subset. Because the groups contain different datasets, this comparison is descriptive.

\begin{table}[!ht]
\centering
\caption{Full-model results by target-dataset anomaly-label availability
(\%). Datasets and sequences refer to the Eval subsets.}
\label{tab:label-availability}
\AppendixTableStyle
\setlength{\tabcolsep}{4pt}
\begin{tabular}{@{}llrrrrrr@{}}
\toprule
Track & Labels & Datasets & Sequences & VUS-PR & AP & VUS-ROC & AUC-ROC \\
\midrule
U & Available & 19 & 346 & 65.94 & 64.57 & 94.07 & 92.70 \\
U & Unavailable & 4 & 4 & 32.45 & 37.55 & 75.58 & 75.77 \\
\midrule
M & Available & 11 & 173 & 51.38 & 52.31 & 87.40 & 86.79 \\
M & Unavailable & 6 & 7 & 23.03 & 28.78 & 76.75 & 78.75 \\
\bottomrule
\end{tabular}
\end{table}

\begin{table}[t]
\centering
\caption{Paired VUS-PR differences on TSB-AD-U/M (percentage points), with
95\% sequence-resampling intervals. Positive values favor HieraNTP.}
\label{tab:paired-ablation}
\AppendixTableStyle
\setlength{\tabcolsep}{5pt}
\begin{tabular}{@{}lcccc@{}}
\toprule
& \multicolumn{2}{c}{TSB-AD-U} & \multicolumn{2}{c}{TSB-AD-M} \\
\cmidrule(lr){2-3}\cmidrule(l){4-5}
Removed component & $\Delta$ & 95\% interval & $\Delta$ & 95\% interval \\
\midrule
Patch variance & $+3.09$ & $[+1.49,+4.72]$ & $+3.03$ & $[+1.49,+4.70]$ \\
Observation-space discrepancy & $+3.02$ & $[+1.24,+4.95]$ & $+2.51$ & $[+0.90,+4.36]$ \\
Temporal summaries & $+6.31$ & $[+4.54,+8.11]$ & $+4.64$ & $[+2.87,+6.56]$ \\
Reference calibration & $+5.28$ & $[+3.32,+7.28]$ & $-1.43$ & $[-4.33,+1.37]$ \\
\bottomrule
\end{tabular}
\end{table}

For each selected ablation, we match its VUS-PR with HieraNTP on the same
evaluation sequence and compute $\Delta$ as full minus ablated performance.
Within each track, 10,000 bootstrap replicates resample complete sequences
with replacement, using identical indices for both configurations. The
2.5th and 97.5th percentiles of the macro-average differences form the
intervals in Table~\ref{tab:paired-ablation}. These quantify sequence-resampling
uncertainty for fixed fitted models. Within-dataset dependence and variability
across training runs remain outside this estimate.
The reference-calibration interval is positive on U and spans zero on M.

\paragraph{Token-angle diagnostic.}
\label{app:token-angle-diagnostic}
We examine the angular response of the token-space statistic in
Equation~\ref{eq:retrieval-evidence}. With random seed 20260924, we sample
one Eval sequence per dataset and one variable per multivariate sequence,
retaining at most 256 token pairs per scale, direction, and alignment.
The sample contains 201,044 pairs from 23 U and 17 M sequences. Target spans
are grouped as normal, anomalous, or mixed according to whether their point
labels are all normal, all anomalous, or contain both. Labels are used only
for this diagnostic. Table~\ref{tab:token-angle} summarizes these groups.

\begin{table}[t]
\centering
\caption{Cosine similarity between predicted and realized tokens in the diagnostic sample.}
\label{tab:token-angle}
\AppendixTableStyle
\setlength{\tabcolsep}{5pt}
\begin{tabular}{@{}llrrrr@{}}
\toprule
Track & Target-span labels & Token pairs & $\cos\theta<0$ (\%) & $\cos\theta<0.5$ (\%) & Median \\
\midrule
U & All & 111,516 & 5.09 & 33.58 & 0.656 \\
  & Normal & 95,723 & 4.75 & 33.11 & 0.659 \\
  & Anomalous & 4,837 & 5.50 & 30.39 & 0.710 \\
  & Mixed & 10,956 & 7.85 & 39.10 & 0.612 \\
\midrule
M & All & 89,528 & 5.96 & 31.05 & 0.687 \\
  & Normal & 76,965 & 5.83 & 30.62 & 0.699 \\
  & Anomalous & 4,593 & 8.14 & 39.32 & 0.633 \\
  & Mixed & 7,970 & 5.93 & 30.41 & 0.639 \\
\bottomrule
\end{tabular}
\end{table}

Negative cosine similarity occurs in 5.09\% of sampled U pairs and 5.96\%
of M pairs, corresponding to prediction--target angles above $90^\circ$.
The tangent statistic measures projected mismatch, weighted by the fitted
scale-specific precision matrix. With identity precision, it equals
$1-\cos^2\theta$ and is nonmonotonic over the full angular range. Using the
fitted matrices, the negative-cosine bin has lower median tangent energy
than the adjacent $[0,0.25)$ cosine bin in eight of 48 track--scale--direction--alignment
groups. No sampled pair has cosine similarity below $-0.9$.

\FloatBarrier

\section{Visualization Details}
\label{app:case-visualization}

\subsection{Main-Text Visualizations}

Figure~\ref{fig:dataset-ablation} averages full-minus-ablated VUS-PR over
each dataset's Eval sequences. Tracks share a symmetric percentage-point color scale.
Within-track clustering uses average linkage and Euclidean distance
over six unstandardized differences.

Figure~\ref{fig:dataset-evidence-usage} aggregates split gain from
dataset-specific appended trees. Source, temporal support, and processing
variant partition the same total, each summing to 100\% before rounding.
Totals include all temporal summaries, alignments, and variable aggregations.
Decoded sources include native-scale and full-resolution evidence, with
Full identifying the latter. Rows use average-linkage clustering of the
five source shares with Euclidean distance. Parentheses give Eval sequence counts.

Figure~\ref{fig:selected-evidence} shows manually selected contiguous 75\%
windows from Exathlon (U) and TAO (M). Each retains the five highest-gain
exported features with nonconstant window responses. Curves are binned,
orange bars mark anomaly labels, and dashed lines mark reference-prefix
boundaries. TAO displays the temporally smoothest variable.

Labels identify source, patch length or full resolution, temporal summary,
grid offset, processing variant, and variable aggregation.
Mean/Max numbers specify window lengths in observations; Current value
uses no temporal aggregation. Variable mean/max aggregates across variables.
A shared grid offset appears once beside the sequence.
Percentages report each feature's share of dataset-extension split gain.
Pool denotes variable aggregation and Cal.\ reference calibration.
Patch mean and Patch variance compare local statistics.
Decoded observation and Decoded target compare predictions with observations
and realized-token reconstructions, respectively.
Heatmaps use full-sequence 2nd--98th percentile ranges per feature;
window VUS-PR uses unbinned scores and labels.

\subsection{Additional Full-Sequence Cases}
\label{app:additional-cases}

\begin{figure}[t]
  \centering
  \includegraphics[width=\linewidth]{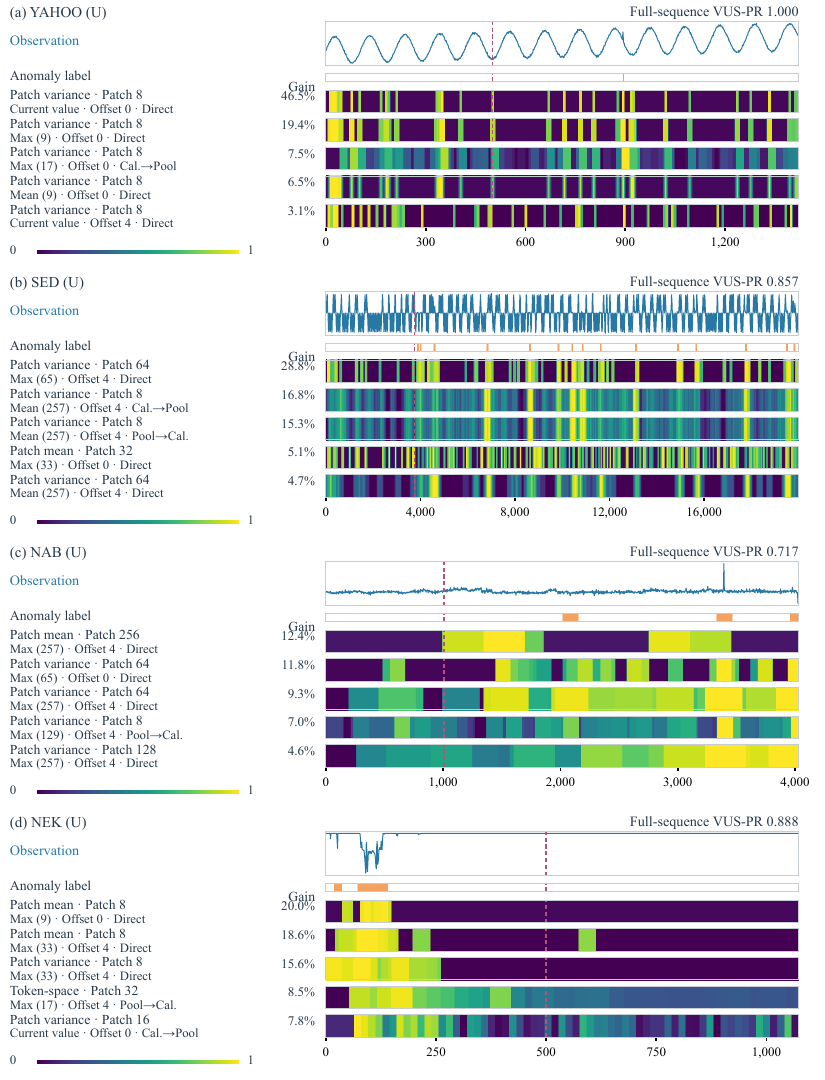}
  \caption{Evidence responses in (a) YAHOO, (b) SED, (c) NAB, and (d) NEK.
  YAHOO emphasizes short-patch variance, whereas SED combines patch lengths
  8 and 64 with sharp maxima and sustained temporal means. NAB's large-patch
  mean evidence forms broad responses around later changes. In NEK,
  short-patch mean evidence concentrates around early deviations, while
  token-space and variance responses extend over different intervals.
  Throughout the gallery, bins show mean responses and any anomaly;
  observations are min--max normalized. Multivariate panels show the smoothest
  variable with appreciable variation, selected without labels. Axes give
  time indices, and dashed lines mark reference-prefix boundaries.}
  \label{fig:additional-variance}
  \label{fig:additional-temporal}
\end{figure}

\begin{figure}[t]
  \centering
  \includegraphics[width=\linewidth]{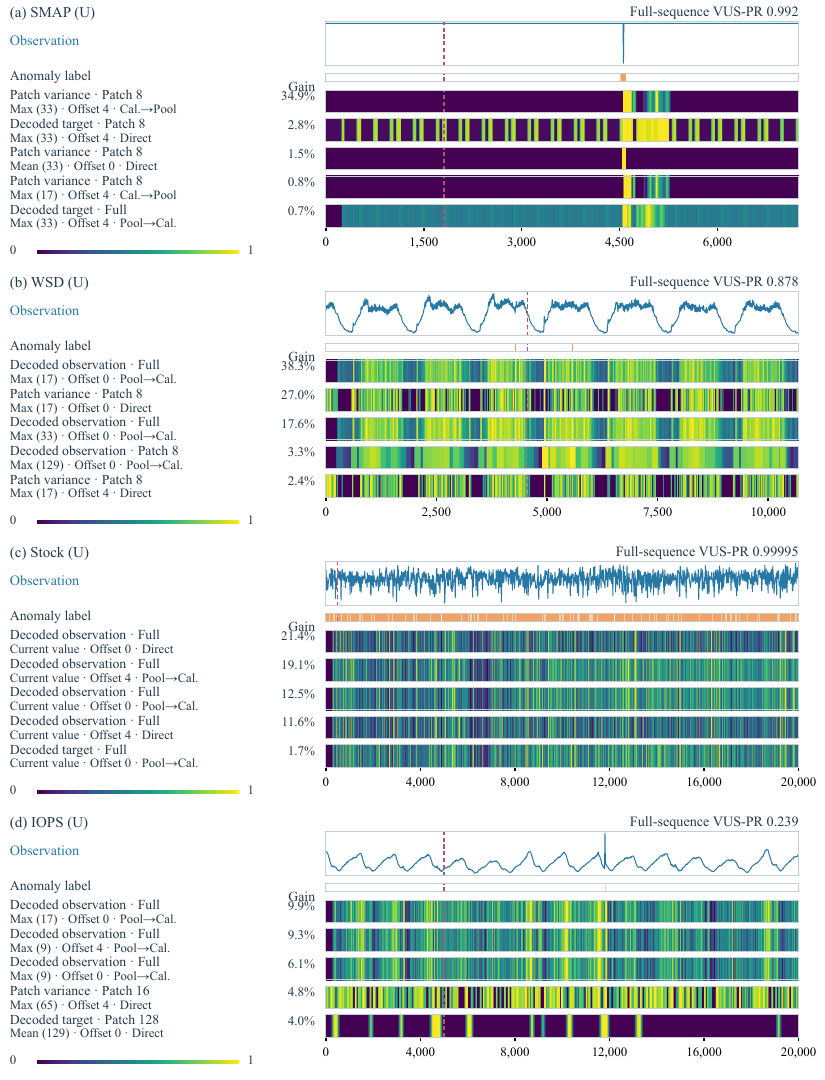}
  \caption{Evidence responses in (a) SMAP, (b) WSD, (c) Stock, and (d) IOPS.
  SMAP's highest-gain variance feature follows the downward
  excursion alongside broader decoded-target responses. WSD combines
  decoded-observation and variance evidence across recurring cycles. Stock
  emphasizes current full-resolution responses. IOPS's decoded summaries
  extend beyond the labeled event.}
  \label{fig:additional-statistic-decoded}
  \label{fig:additional-decoded}
\end{figure}

\begin{figure}[t]
  \centering
  \includegraphics[width=\linewidth]{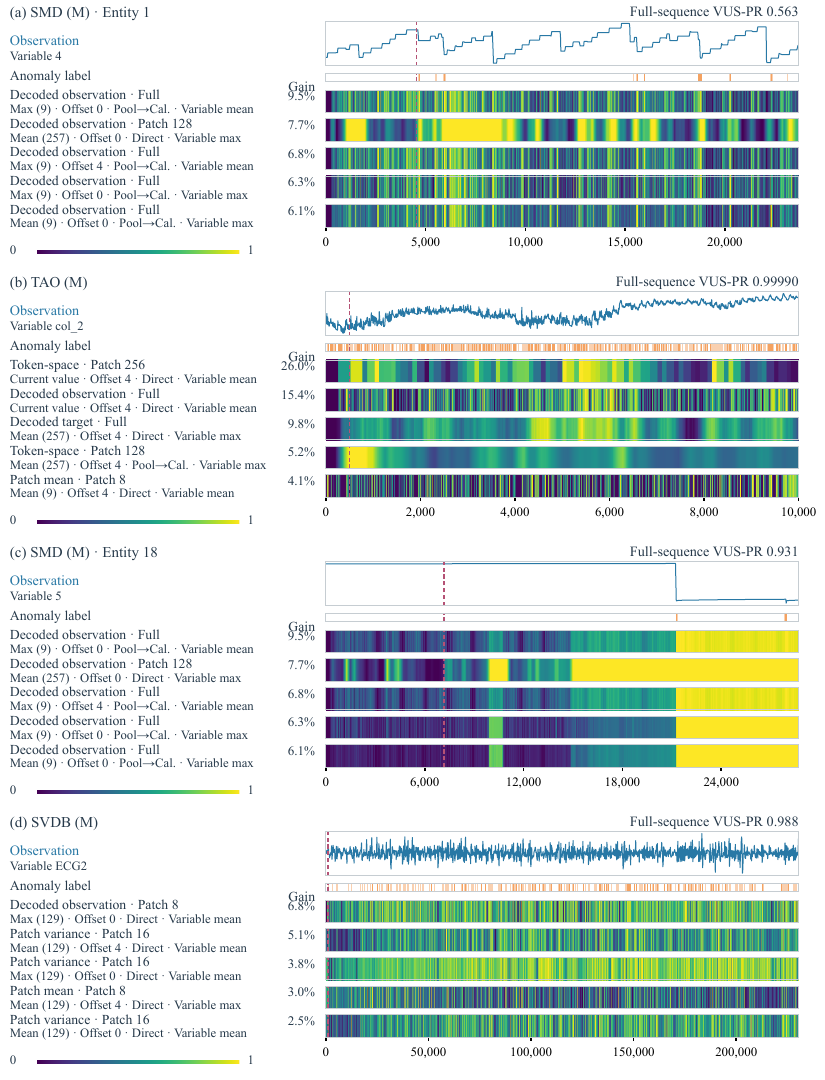}
  \caption{Multivariate evidence in (a) SMD entity 1, (b) TAO entity 7,
  (c) SMD entity 18, and (d) SVDB entity 14. The same SMD features respond
  to recurring changes in (a) and the later level drop in (c). TAO combines
  slowly varying token-space responses with faster decoded-observation and
  temporally averaged decoded-target evidence. SVDB combines decoded,
  variance, and mean evidence with distinct temporal response patterns.}
  \label{fig:additional-smd-tao}
  \label{fig:additional-multivariate}
\end{figure}

Figures~\ref{fig:additional-variance}--\ref{fig:additional-multivariate}
extend Section~\ref{sec:evidence-usage} with eight univariate and four
multivariate sequences. Stock, TAO, SMD entity 18, and SVDB were selected
for high full-sequence VUS-PR. Each panel displays five nonconstant exported
forward-direction features, ranked by split gain in the dataset-specific
extension. Gain shares describe the fitted extension across its training
samples. Reported VUS-PR uses the bidirectional detector's original-resolution
scores and labels over the complete sequence.

\FloatBarrier

\section{Limitations}
\label{app:limitations}

HieraNTP relies on anomaly labels for evidence selection. With normal-only
target training data, the shared detector learned from other datasets may
inadequately capture the target anomaly definition. The current XGBoost
detector also relies on empirically designed risk evidence, motivating future
work on adaptive evidence representations learned with deep models.

Multivariate detection aggregates evidence across variables without explicitly
modeling their dependencies, limiting sensitivity to anomalies in
inter-variable relationships. The pipeline is currently offline, and removing
reverse-direction evidence reduces performance in our ablations. Adapting it
to real-time detection may therefore incur an accuracy cost.

\end{document}